\documentclass[letterpaper, 10 pt, conference]{ieeeconf} 
\IEEEoverridecommandlockouts
\usepackage{url}
\usepackage{ulem}
\usepackage{graphicx}
\usepackage[dvipsnames]{xcolor}
\usepackage{multirow}
\usepackage{booktabs}
\usepackage{arydshln}
\usepackage{hyperref}
\usepackage{amsmath, amssymb}
\usepackage{makecell}
\usepackage{colortbl}
\usepackage{placeins}  
\usepackage{fontawesome5}
\usepackage[noadjust]{cite}
\usepackage{microtype}

\DeclareMathOperator*{\argmax}{arg\,max}

\usepackage[caption=false, font=footnotesize]{subfig}

\usepackage{eso-pic}
\usepackage{url}
\AddToShipoutPicture*{%
     \AtTextUpperLeft{%
         \put(-3.5,10){
           \begin{minipage}{\textwidth}
              \scriptsize
              \MakeUppercase{IEEE/RSJ International Conference on Intelligent Robots and Systems. Preprint version. Accepted June 2025.} 
              \\
              Preprint version; final version available at \url{https://doi.org/10.1109/IROS60139.2025.11247445}
           \end{minipage}}%
     }%
}

\hypersetup{
    colorlinks=true,
    urlcolor=blue,
    linkcolor=blue,
    citecolor=blue
}

\definecolor{blue10}{rgb}{0.3, 0.5, 0.8}
\newcommand{\bluehref}[2]{\url{#1}}

\newcommand{\etal}{et al.}
\definecolor{random}{rgb}{0.1216, 0.4667, 0.7059}
\definecolor{mixvpr}{rgb}{1.0000, 0.4980, 0.0549}
\definecolor{cosplace}{rgb}{0.1725, 0.6275, 0.1725}
\definecolor{netvlad}{rgb}{0.8392, 0.1529, 0.1569}
\definecolor{anyloc}{rgb}{0.5804, 0.4039, 0.7412}
\definecolor{megaloc}{rgb}{0.498, 0.498, 0.498}
\definecolor{crica}{rgb}{0.082, 0.745, 0.812}
\definecolor{superpointbrute}{rgb}{0.5490, 0.3373, 0.2941}
\definecolor{hierarchical}{rgb}{0.8902, 0.4667, 0.7608}
\newcommand{\myline}[1]{\textcolor{#1}{\raisebox{0.5ex}{\rule{0.4cm}{3.5pt}}}}

\title{\LARGE \bf
Image-Based Relocalization and Alignment for Long-Term Monitoring of Dynamic Underwater Environments \faWater
}

\author{Beverley Gorry \qquad \quad Tobias Fischer \qquad \quad Michael Milford \qquad \quad Alejandro Fontan %
\thanks{All authors are with the QUT Centre for Robotics, School of Electrical Engineering and Robotics, Queensland University of Technology, Brisbane, QLD 4000, Australia. This research was partially supported by funding from ARC Laureate Fellowship FL210100156 to MM and ARC DECRA Fellowship DE240100149 to TF, as well as through the Reef Restoration and Adaptation Program, a partnership between the Australian Government’s Reef Trust and the Great Barrier Reef Foundation. The authors acknowledge continued support from the Queensland University of Technology (QUT) through the Centre for Robotics. Corresponding author email:
        {\tt\small beverley.gorry@hdr.qut.edu.au}}%
}

\begin{document}

\maketitle
\thispagestyle{empty}
\pagestyle{empty}

\begin{abstract}
Effective monitoring of underwater ecosystems is crucial for tracking environmental changes, guiding conservation efforts, and ensuring long-term ecosystem health. However, automating underwater ecosystem management with robotic platforms remains challenging due to the complexities of underwater imagery, which pose significant difficulties for traditional visual localization methods. We propose an integrated pipeline that combines Visual Place Recognition (VPR), feature matching, and image segmentation on images extracted from video sequences. This method enables robust identification of revisited areas, estimation of rigid transformations, and downstream analysis of ecosystem changes.  
Furthermore, we introduce the SQUIDLE+ VPR Benchmark—the first large-scale underwater VPR benchmark designed to leverage an extensive collection of unstructured data from multiple robotic platforms, spanning time intervals from days to years. The dataset encompasses diverse trajectories with varying overlap and diverse seafloor types captured under different environmental conditions, including differences in depth, lighting, and turbidity. Our code is available at: \bluehref{https://github.com/bev-gorry/underloc}{UnderLoc}.
\end{abstract}
\section{Introduction}
\label{sec:introduction}
Underwater ecosystems like coral reefs, seagrass beds, and kelp forests support marine biodiversity, fisheries, and coastal protection. These underwater habitats are vital but face growing threats from climate change, pollution, and human activities such as overfishing and coastal expansion~\cite{hoegh2010impact,hughes2017coral}. Long-term monitoring of these habitats is essential to detect and quantify changes in the distribution and abundance of different species, helping researchers better understand the impacts of human activities and 
improving the long-term health and resilience of these critical ecosystems~\cite{madin2019emerging}.

Due to the cost and limitations of diver-based monitoring, autonomous underwater vehicles offer a scalable alternative~\cite{dunbabin2012robots}. Equipped with AI and computer vision, they enable efficient, non-intrusive monitoring with advances in navigation~\cite{zhang2023autonomous}, 3D mapping~\cite{iscar2017multi,wang2023real}, and image segmentation~\cite{raine2024image} for high-resolution data collection.

\begin{figure}[t]
\centering
\includegraphics[width=\linewidth]{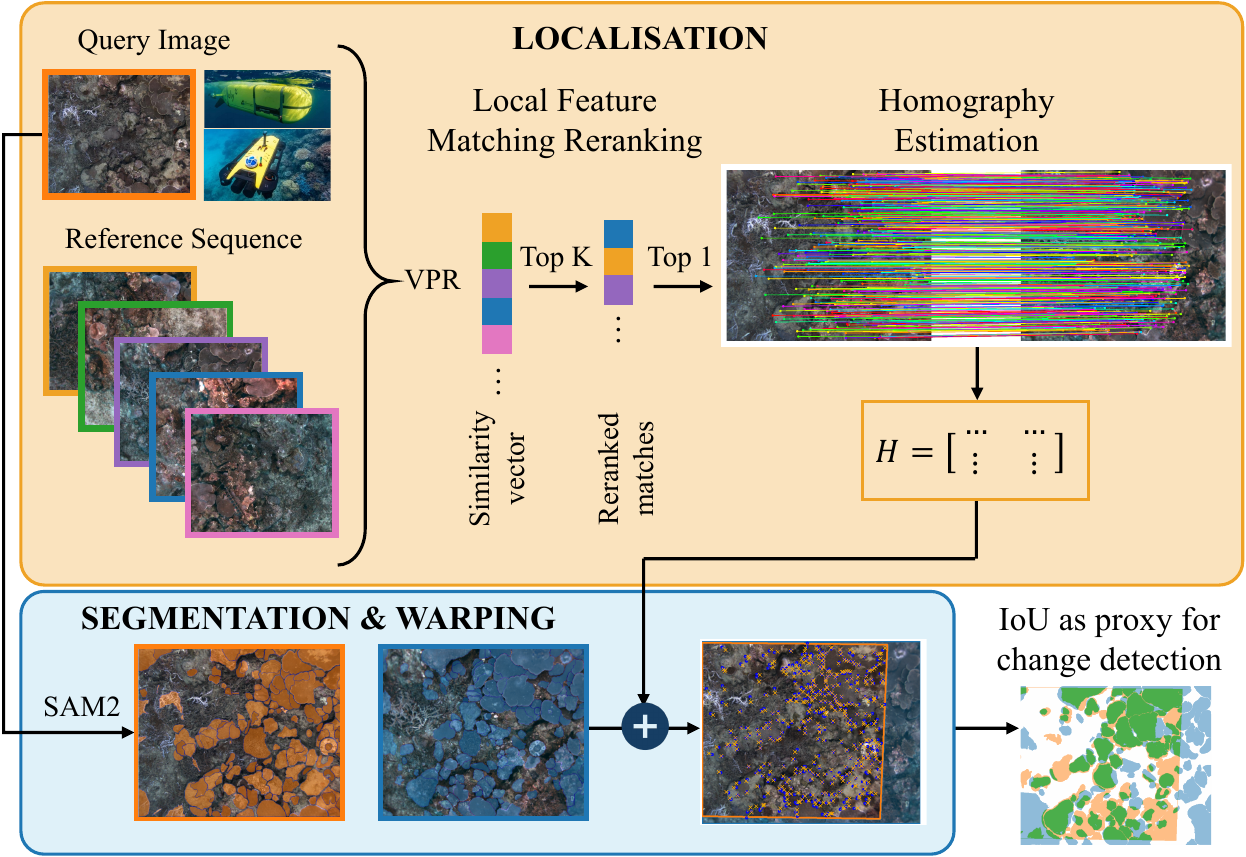}
\caption{\textbf{Overview of our image-based relocalization and alignment method for underwater ecosystem monitoring.} We use hierarchical Visual Place Recognition (VPR) techniques to robustly identify common locations from images captured across multi-year timescales using video from freely navigating robots. These images exhibit significant variations in environmental conditions, such as lighting, turbidity, and depth.  We establish correspondences between image keypoints to estimate a rigid transformation between the images, which we then use to register segmentation masks in a common pixel space. Finally, we apply an intersection over union (IoU) metric to detect ecosystem changes over time.}
\vspace*{-0.3cm}
\label{fig:teaser_fig}
\end{figure}

Reliable multi-year change detection enables long-term monitoring of these dynamic and fragile ecosystems, requiring highly accurate registration to capture centimeter-level variations in habitats such as coral reefs~\cite{delaunoy2008towards}. In terrestrial and aerial monitoring scenarios, change detection benefits from GPS location priors, well-defined landmarks, and many decades of robotic vision techniques mostly tailored for these scenarios. However, these methods do not directly translate to underwater surveys, which bring unique challenges such as limited visibility, optical distortions, turbidity, caustics, and non-linear light attenuation across depths. 

Early work by Eustice et al.~on large-scale hull mapping laid the groundwork for underwater Simultaneous Localization and Mapping (SLAM) techniques, using both vision-based and sonar-based methods for infrastructure inspection~\cite{eustice2005visually,eustice2005exactly}. These efforts targeted applications like ship hull inspection~\cite{kim2009pose}, oil rig monitoring, and pipeline surveying~\cite{fairfield2007real}, which, while distinct from ecosystem monitoring, share challenges related to long-term feature tracking~\cite{rahman2022svin2}, sensor degradation, and environmental variability~\cite{delaunoy2008towards}.

In this paper, we focus on long-term environmental monitoring enabled by recent advances in visual place recognition (VPR) and semantic segmentation. VPR enables autonomous systems to recognize previously visited locations based on a set of reference images. While VPR has the potential for accurate, repeatable localization, we show that current state-of-the-art techniques are challenged by new, difficult datasets in the underwater domain, which has received very limited attention thus far. Our approach requires only a set of pre-collected images as a reference -- there is no need for SLAM, Structure-from-Motion (SfM), or geometric mapping. This makes it simpler and more scalable for long-term monitoring.

As depicted in Figure~\ref{fig:teaser_fig}, our system enables cost-effective and scalable underwater ecosystem monitoring by registering locations across multi-year timescales using monocular video from autonomous marine robots. It identifies revisited areas, estimates rigid transformations, and analyzes visual changes, even under varying trajectory overlap. By integrating VPR with feature matching and image warping, our approach enhances spatial consistency and registration accuracy over time. While our method builds upon existing techniques, we are the first to combine them in this manner, enabling critical monitoring capabilities that are valuable to ecologists.

Finally, we introduce the SQUIDLE+ VPR Benchmark, the first large-scale benchmark for evaluating underwater VPR, leveraging publicly available data from SQUIDLE+. In extensive experiments, we benchmark state-of-the-art VPR techniques for underwater environments, advancing precise registration for long-term ecological monitoring and improving the reliability, scalability, and effectiveness of marine surveys for scientific research and marine conservation.

\begin{figure*}[t]
    \scriptsize
    \includegraphics[width=0.98\linewidth]{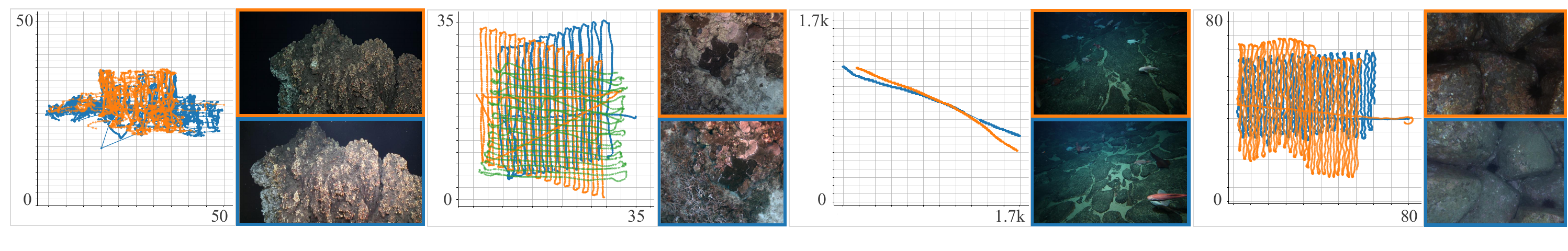} \\
    \hspace*{1cm}(a) Eiffel Tower {\color{random}2018}-{\color{mixvpr}2020} \hspace*{1.5cm} (b) Okinawa {\color{cosplace}2016}-{\color{random}2017}-{\color{mixvpr}2018}  \hspace*{.8cm} (c) Tasman Fracture 2018/12/{\color{random}04}-{\color{mixvpr}06} \hspace*{1.1cm} (d) St Helens {\color{random}2011}-{\color{mixvpr}2013}
    \caption{\textbf{GPS trajectories (in meters) and environmental differences across underwater datasets.} The visualized trajectories depict query sequences ({\color{mixvpr}orange}) and database sequences ({\color{random}blue}) for four underwater datasets: (a) the \textit{Eiffel Tower} (2018–2020) from the Mid-Atlantic Ridge hydrothermal vent, (b) \textit{Okinawa} (2016–2017–2018), capturing mesophotic coral reef environments before and after Typhoon Trami,  (c) the \textit{Tasman Fracture} (2018), showcasing deep-sea benthic habitats, and  (d) \textit{St Helens} (2011–2013), featuring images recorded during a transition movement in shallow barren zones. To analyze the unique overlap between sequences covering the same area but following different trajectories, we include a third trajectory in the Okinawa dataset ({\color{cosplace}green}). Accompanying RGB image pairs for each trajectory illustrate corresponding locations, highlighting the significant appearance variations that challenge automated VPR systems. Notably, even correctly matched locations exhibit substantial visual differences due to variations in viewpoint, lighting conditions, and actual ecosystem changes over time, further emphasizing the complexity of long-term visual place recognition in underwater environments.}
    \label{fig:gps_trajectories}
    \vspace*{-0.2cm}
\end{figure*}

\section{Related Works}
\label{sec:related}
We review image registration, 3D mapping, and VPR as key components of long-term underwater ecosystem monitoring. Traditional image registration (Section~\ref{subsec:imagereg}) relies on GPS or multi-sensor SLAM for coarse localization before alignment refinement, while our approach uses VPR to efficiently retrieve images from monocular video for precise registration. In 3D mapping, existing methods propagate semantic information within a single trajectory, whereas we extend segmentation across multi-year datasets under varying conditions (Section~\ref{subsec:3dmapping}). Underwater VPR remains underexplored, with most techniques designed for structured terrestrial environments (Section~\ref{subsec:vpr}). Finally, we review benchmarking in underwater localization (Section~\ref{subsec:benchmarking}).

\subsection{Image Registration for Change Detection}
\label{subsec:imagereg}
Delaunoy \etal~\cite{delaunoy2008towards} used SURF features to register and detect changes in images taken 10 months apart in a small coral reef scene. Similarly, Williams \etal~\cite{williams2010repeated} applied a stereo-based SLAM technique with SIFT features to co-register multiple 3D image maps collected by an autonomous underwater vehicle (AUV) during a grid survey. Their approach compared the same area over 12 hours, estimating the AUV’s trajectory across overlapping grid-based surveys.

Bryson \etal~\cite{bryson2013automated} advanced multi-year repeat survey imagery processing and precision registration for monitoring long-term changes in benthic marine habitats. Post-processing of stereo imagery and navigation data through SLAM and 3D reconstruction generated a photo-textured model of the seafloor, orthographically projected into a geo-referenced mosaic. Their approach included mutual information optimization to improve robustness against variations in color and brightness across years. 

These studies relied on GPS or multi-sensor SLAM for coarse localization before applying image registration methods. In contrast, our method efficiently retrieves image cues from the same location using videos from an uncalibrated monocular camera. This enables accurate image registration as required for monitoring applications.

\subsection{3D Mapping of Coral Reefs}  
\label{subsec:3dmapping}
Recent advances in underwater 3D mapping have improved coral reef reconstruction and semantic interpretation. Sauder \etal~\cite{Sauder:Semantic3DMappingCoralReefsDL:2024} proposed a pipeline to propagate semantic segmentation from images to 3D point clouds estimated via SfM from single-trajectory video streams. Sethuraman \etal~\cite{sethuraman2023waternerf} leveraged neural radiance fields (NeRFs) for physics-informed novel view synthesis and image restoration, addressing water column effects like attenuation and backscattering. Wang \etal~\cite{wang2023real} combined visual-inertial odometry with real-time 3D reconstruction to generate dense maps on resource-constrained AUVs, achieving results comparable to offline methods. Song \etal~\cite{song2024turtlmap} introduced TURTLMap, designed for real-time localization and mapping in low-texture underwater regions.

Unlike these approaches that typically focus on single-time mapping, require controlled camera paths, or address specific visibility challenges, our method enables propagation of semantic segmentation across images captured in different years under varying conditions with arbitrary trajectory overlaps. Our pipeline prioritizes image-to-image registration for direct ecological comparison using only monocular video input, making it more accessible for practical long-term monitoring applications.

\subsection{Visual Place Recognition}
\label{subsec:vpr}
Despite its potential, underwater VPR remains largely underexplored and is rarely applied in practical settings. Most VPR pipelines~\cite{Arandjelović:NetVLAD:2018, Hausler:PatchNetVLAD:2021, Berton:Cosplace:2022, lu2024cricavpr, izquierdo2024optimal, ali2023mixvpr} are developed and tested in structured environments, with limited application to unstructured settings—and almost none for underwater imagery. A thorough review of terrestrial place recognition techniques is beyond the scope of this paper, but excellent surveys exist for the interested reader~\cite{Zhang:VPRSurveyDeepLearningPerspective:2021,Masone:SurveyDeepVPR:2021,Garg:WhereIsYourPlace:2021,Schubert:VPRTutorial:2024}.

Of interest for this paper are AnyLoc~\cite{Keetha:AnyLoc:2024}, which is one of the few VPR techniques evaluated on the Eiffel Tower underwater dataset~\cite{Boittiaux:EiffelTower:2023}, and 
MegaLoc~\cite{berton2025megalocretrievalplace}, a versatile image retrieval model that achieved state-of-the-art performance in VPR, visual localization, and landmark retrieval by leveraging a diverse set of data and training techniques.

Place recognition is often used in a hierarchical pipeline, where the top $K$ retrieved database images are reranked using more computationally intensive local feature matching. A widely used example is hloc~\cite{sarlin2019coarse}, which uses NetVLAD~\cite{Arandjelović:NetVLAD:2018} for image retrieval and SuperPoint~\cite{detone2018superpoint} with SuperGlue~\cite{Sarlin:SuperGlue:2020} pre-trained on outdoor scenes for local matching.

Several works specifically address place recognition in underwater environments. Maldonado-Ramirez \etal~\cite{Maldonado:LearningAdHocVPRUnderwater:2019} propose an unsupervised VPR method using a convolutional autoencoder to learn compact representations from salient landmarks detected by a visual attention algorithm. Burguera \etal~\cite{Burguera:VisualLoopDetectionUnderwaterDL:2020} introduce a deep network for fast, robust underwater loop detection with clustered SIFT features and unsupervised training. While these methods focus on underwater VPR and evaluate self-recorded robotic sequences, our approach is tested on large, publicly available benchmarks covering diverse sequences, robotic platforms, and camera setups, aligning more closely with general VPR evaluation.

\subsection{Underwater Localization Benchmarking}
\label{subsec:benchmarking}
Boittiaux \etal~\cite{Boittiaux:EiffelTower:2023} introduced the Eiffel Tower dataset, a deep-sea dataset for long-term visual localization. Ferrera \etal~\cite{ferrera2019aqualoc} presented AQUALOC to support visual–inertial–pressure SLAM for underwater vehicles, providing offline trajectories from SfM for comparison with real-time localization methods.

Joshi \etal~\cite{joshi_gopro_icra_2022} proposed a Visual-Inertial SLAM pipeline, evaluated on a wreck off South Carolina and in Florida’s caverns and caves. Singh \etal~\cite{singh2024online} introduced a dataset for benchmarking refractive camera model estimation, featuring a time-synchronized 5-camera/IMU setup on a remotely operated vehicle (ROV) in a controlled pool environment.

Angelakis \etal~\cite{angelakis11using} used animal-borne video and movement data from Australian sea lions to map benthic habitats, addressing challenges in vessel-based surveys, which are costly, time-consuming, weather-dependent, and impractical for deep-sea environments. Joshi \etal~\cite{joshi20243d} presented a 3D water quality mapping system for shallow waters using a BlueROV2 with GPS and a water quality sensor, enabling location correction by resurfacing when errors occur.

In this paper, we make a step towards standardized benchmarking of VPR methods in underwater scenarios, akin to efforts in terrestrial VPR~\cite{Berton_CVPR_2022_benchmark,Zaffar:VPRBench:2021}. Specifically, we propose to leverage sequences freely available on \bluehref{https://squidle.org/}{SQUIDLE+}. SQUIDLE+ hosts the largest repository of openly accessible georeferenced marine images, enabling the construction of datasets tailored to specific scientific questions and sourced from various platforms, campaigns, and globally distributed deployments. By leveraging this vast data source, our work establishes the foundation for a dedicated underwater VPR benchmark, facilitating the development and rigorous evaluation of VPR pipelines.

\begin{table}[t]
\setlength{\tabcolsep}{3pt}
\newcommand{\ang}{90}
\caption{\textbf{SQUIDLE+ VPR Benchmark} originates from various robotic platforms following trajectories with arbitrary overlap, spanning time differences ranging from days to years. The data includes diverse seafloor types captured under varying scene conditions (e.g., depth, lighting, turbidity).}
\label{tab:datasets}
\centering
\resizebox{\columnwidth}{!}{ %
\begin{tabular}{lcccccccccc}%
\arrayrulecolor{blue10}
&\rotatebox{0}{\textbf{Dataset}} & \rotatebox{\ang}{\textbf{Year}} & \rotatebox{\ang}{\textbf{\# Images (k)}} & \rotatebox{\ang}{\textbf{Dur. (h)}} & \rotatebox{\ang}{\textbf{Dist. (km)}} & \rotatebox{\ang}{\textbf{Avg. Depth (m)}} & \rotatebox{\ang}{\textbf{Seafloor}} & \rotatebox{\ang}{\textbf{Platform}} & \rotatebox{\ang}{\textbf{Loc. Radius (m)}} \\ \midrule
\multirow{8}{*}{\rotatebox{90}{\textbf{SQUIDLE+}}}&\multirow{3}{*}{Okinawa} & 2018 & 2.8 & 01:16 & 0.7 & 37 & \multirow{3}{*}{\makecell{Mesophotic \\ coral}} & \multirow{3}{*}{\makecell{SOTON/UTOK \\OPLab AUV}} & \multirow{3}{*}{5.0}\\
&& 2017 & 3.0 & 02:04 & 0.7 & 36 \\
&& 2016 & 2.1 & 00:58 & 0.5 & 38 \\
\cmidrule(lr){2-10}
&\multirow{2}{*}{\makecell[l]{Tasman\\Fracture}} & 2018/12/04 & 1.2 & 00:57 & 2.3 & 885 & \multirow{2}{*}{\makecell{Seamount \\ habitats}} &\multirow{2}{*}{\makecell{CSIRO O\&A MRITC \\Towed Stereo Camera}} & \multirow{2}{*}{10.0}\\
& & 2018/12/06 & 1.3 & 00:56 & 2.3 & 885 \\
\cmidrule(lr){2-10}
&\multirow{3}{*}{\makecell{St Helens}} & 2013 & 6.5 & 01:48 & 2.9 &  12 & \multirow{3}{*}{\makecell{Shallow \\ barren zones}} &\multirow{4}{*}{\makecell{IMOS AUV \\ Sirius}} & \multirow{3}{*}{3.0}\\
&& 2011 & 5.5 & 01:30 & 1.9 & 12 &&\\
&& 2009 & 6.4 & 01:47 & 2.7 & 10 &&\\
\midrule
&\multirow{4}{*}{\makecell[l]{Eiffel Tower \\(Mid-Atlantic\\Ridge)}} & 2020 & 3.9 & 03:18 & 0.9 & 1.7k& \multirow{4}{*}{\makecell{Hydrothermal \\ vent  edifice}} &
\multirow{4}{*}{\makecell{EMSO-Azores \\ROV}} & \multirow{4}{*}{0.5} \\
&& 2018 & 5.4 & 17:19 & 1.0 & 1.7k &&\\
&& 2016 & 3.6 & 03:24 & 0.9 & 1.7k &&\\
&& 2015 & 4.9 & 01:26 & 0.3 & 1.7k &&\\ 
\bottomrule
\end{tabular}
}
\end{table}

\section{Technical Approach}

This section details our approach to enable change detection in underwater scenes using video streams from uncalibrated monocular cameras. Given two undistorted sequential image sets, $Q$ and $D$, from a video stream, we first find the top $K$ matches in $D$ for each $Q_i$ using global descriptors obtained via state-of-the-art VPR techniques~\cite{ali2023mixvpr,Berton:Cosplace:2022,Arandjelović:NetVLAD:2018,Keetha:AnyLoc:2024,berton2025megalocretrievalplace,lu2024cricavpr}. Using LightGlue~\cite{Lindenberger:LightGlue:2023}, we rerank these matches via the inlier count, and filter out image pairs that have a large reprojection error to avoid false matches. We then extract semantic segmentation masks from the matched images, and warp the images using the homography matrix. We use the intersection-over-union (IoU) to approximate location similarity.

\subsection{Visual Place Recognition – Global Retrieval Stage}
\label{subsec:VPR}

We extract image descriptors using standard VPR techniques and compute the similarity matrix $S \in \mathbb{R}^{|D| \times |Q|}$ where $S_{j,i}$ represents the similarity between database image $D_j$ and query image $Q_i$ using the L2 norm. For each query image $Q_i$, we select the top $K$ database images with the highest similarity scores, forming the set of initial matches:
\begin{equation}
\mathcal{M}_{K,i} = \{(Q_i, D_k) \mid D_k \in \operatorname{TopK}(\{D_j : S_{j,i}\}_{j=1}^{|D|}, K) \},
\end{equation}
where \( \operatorname{TopK}(\cdot, K) \) denotes the function that selects the \( K \) most similar matches.

\subsection{Keypoint Correspondences – Local Refinement Stage}

Using default parameters, we refine matches by using LightGlue to establish keypoint correspondences between SuperPoint features for each candidate pair in $\mathcal{M}_{K,i}$. For a query-database pair $(Q_i, D_k)$, let:
\begin{equation}
\mathcal{C}_{i,k} = \{(\boldsymbol{p}_{q}, \boldsymbol{p}_{d}) \mid \boldsymbol{p}_{q} \in \mathbb{R}^2, \boldsymbol{p}_{d} \in \mathbb{R}^2 \}
\end{equation}
represent the set of keypoint correspondences between query image $Q_i$ and database image $D_k$, where $\boldsymbol{p}_{q}$ and $\boldsymbol{p}_{d}$ are corresponding keypoints in image coordinates. We refer to a keypoint correspondence retained by LightGlue under its default matching criteria as an \textit{inlier}.

We then re-rank the matches based on the inlier count as a new similarity measure, selecting the best match per query:
\begin{equation}
\mathcal{M}_{1,i} = \{(Q_i, D^*) \mid D^* = \argmax_{D_k \in \mathcal{M}_{K,i}} |\mathcal{C}_{i,k}| \}
\end{equation}

\subsection{Homography Estimation and Outlier Filtering}
\label{subsec:matching}
From the image correspondences $\mathcal{C}_{i,k^*}$ in the best match $(Q_i, D_{k^*}) \in \mathcal{M}_{1,i}$, we estimate the homography matrix $H \in \mathbb{R}^{3\times3}$ and compute the bidirectional reprojection error. The reprojection error $e_r$ is computed as the average of the root-mean-square error (RMSE) in both transformation directions:
\begin{align}
    e_r = \frac{1}{2} \Bigg(
    & \sqrt{\frac{1}{|\,\mathcal{C}_{i,k^*}|} \sum_{(\boldsymbol{p}_q, \boldsymbol{p}_d) \in \mathcal{C}_{i,k^*}} \left\| \boldsymbol{p}_q - H \boldsymbol{p}_d \right\|^2}  \notag \\
    & + \sqrt{\frac{1}{|\,\mathcal{C}_{i,k^*}|} \sum_{(\boldsymbol{p}_q, \boldsymbol{p}_d) \in \mathcal{C}_{i,k^*}} \left\| H^{-1} \boldsymbol{p}_q - \boldsymbol{p}_d \right\|^2} \Bigg),
\end{align}
where $|\,\mathcal{C}_{i,k^*}|$ denotes the number of matched keypoints. To account for scale variations, such as when the query image is a zoomed-in version of the database image or vice versa, we compute bidirectional reprojection errors and average them to obtain a symmetric similarity measure.

Finally, we discard image pairs with reprojection error exceeding $\chi=10$ pixels to retain only geometrically consistent matches:
\begin{equation}
\mathcal{M}_\chi = \{(Q_i, D_{k^*}) \mid e_r(Q_i, D_{k^*}) \leq \chi \}.
\end{equation}
This threshold balances tolerance for minor viewpoint variations with rejection of geometrically inconsistent correspondences.

\subsection{2D Warping of Segmentation Masks}
\label{subsec:segmentation}
To simulate a potential change detection method for visual monitoring on the registered image pairs \( \mathcal{M}_\chi \), we automatically extract segmentation masks for each image using Segment-Anything 2 (SAM2)~\cite{Ravi:SAM2:2024}. For simplicity in our experiments, the prompt-free automatic mask generator is used and all segmented instances within an image are merged into a single unified mask. We then use the homography matrix to warp the masks into a common image space, enabling pixel-level comparison using pixel intersection over union (IoU) as a similarity proxy.
\section{Experimental Setup}
\label{sec:experimentalsetup}

\begin{figure*}[t]
\centering
\resizebox{0.9\linewidth}{!}{
\begin{tabular}{ccccc}
\multicolumn{5}{c}{{\myline{random}
} RANDOM {\myline{mixvpr}} MIXVPR {\myline{cosplace}} COSPLACE {\myline{netvlad}} NETVLAD {\myline{anyloc}} ANYLOC {\myline{megaloc}} MEGALOC {\myline{crica}} CRICAVPR {\myline{superpointbrute}} SUPERPOINT-B} \\
\end{tabular}}
\resizebox{\linewidth}{!}{ %
\begin{tabular}{ccccc}
\includegraphics[width=0.25\linewidth]{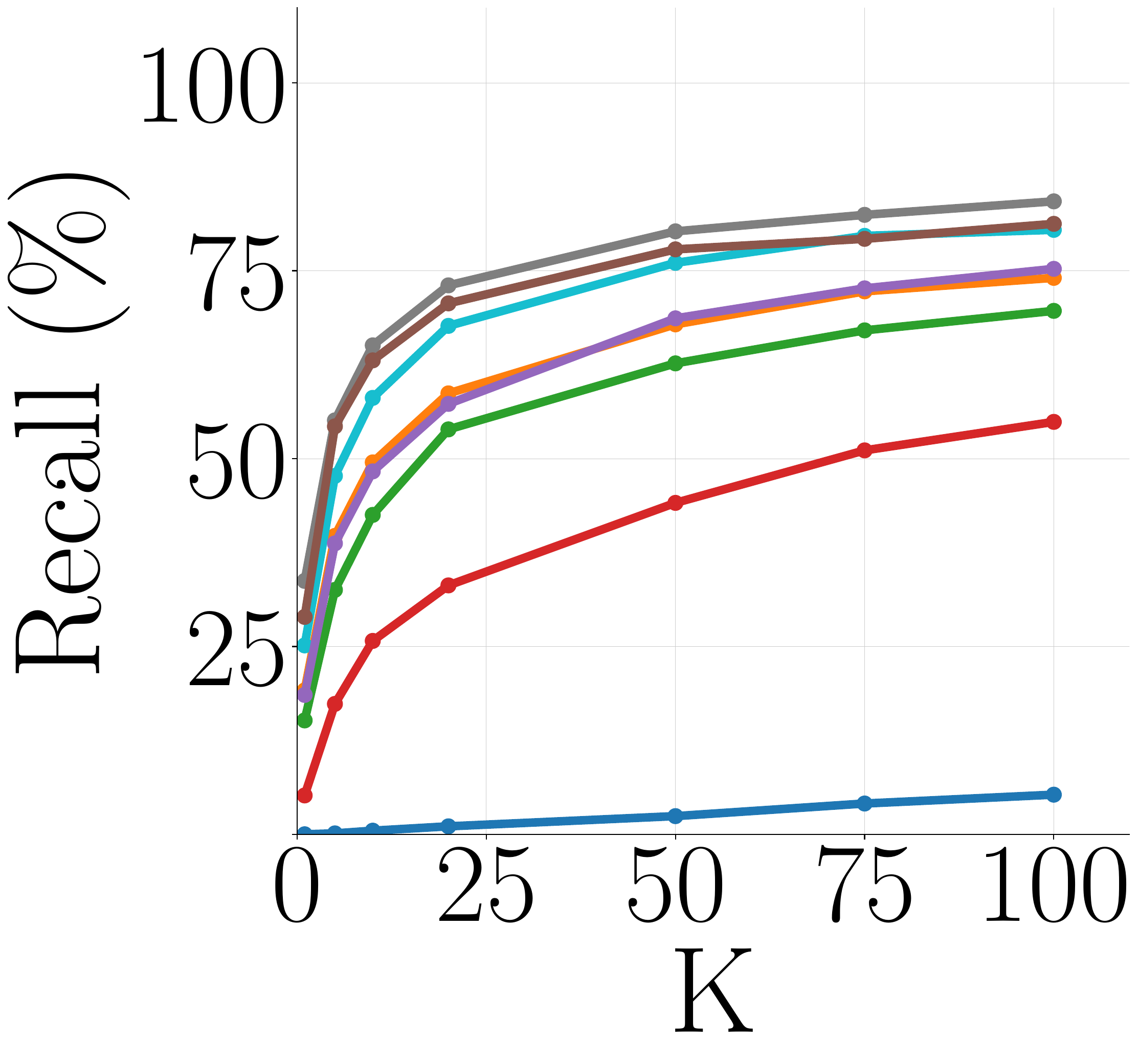} 
&
\includegraphics[width=0.25\linewidth]{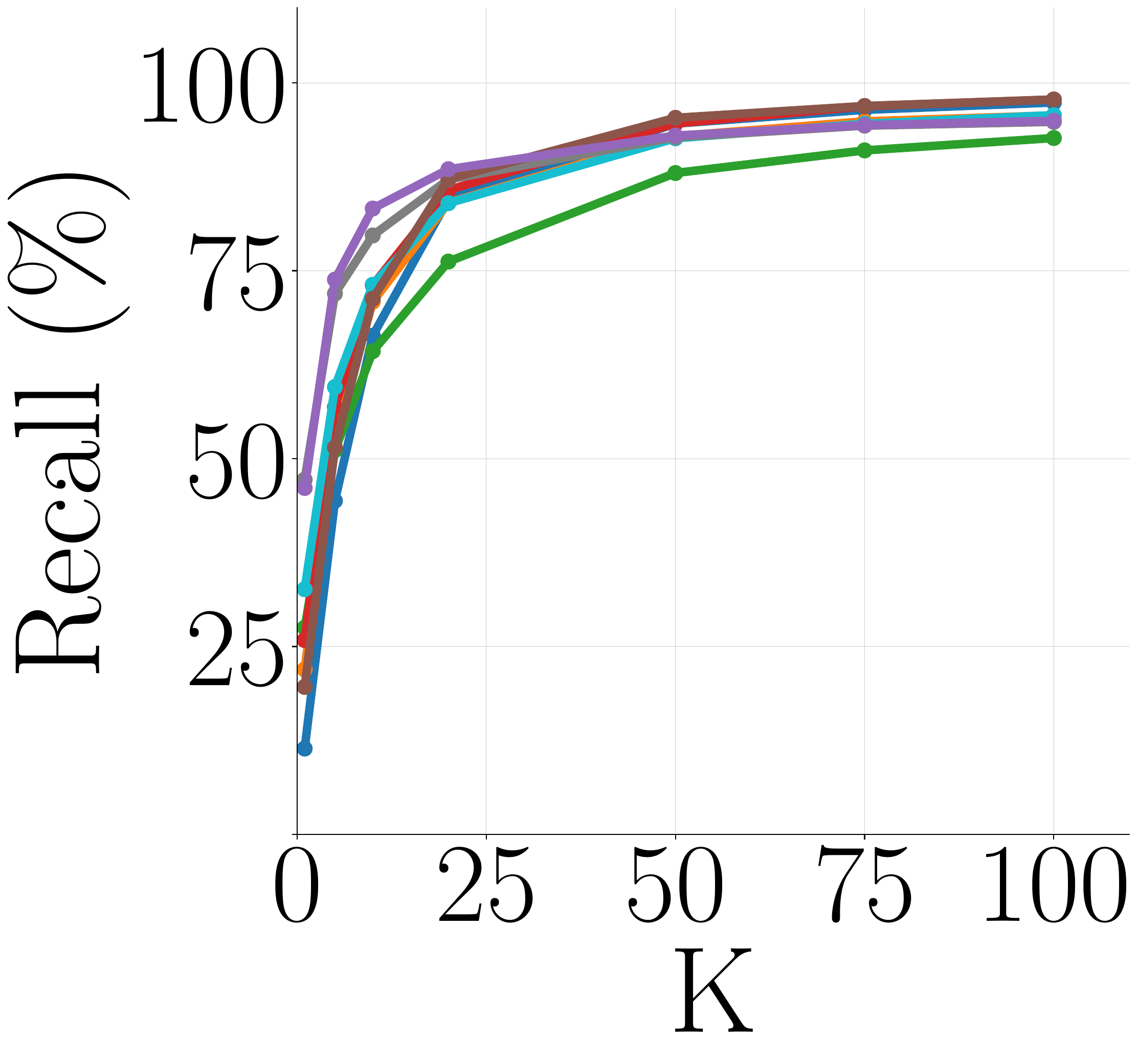}
&
\includegraphics[width=0.25\linewidth]{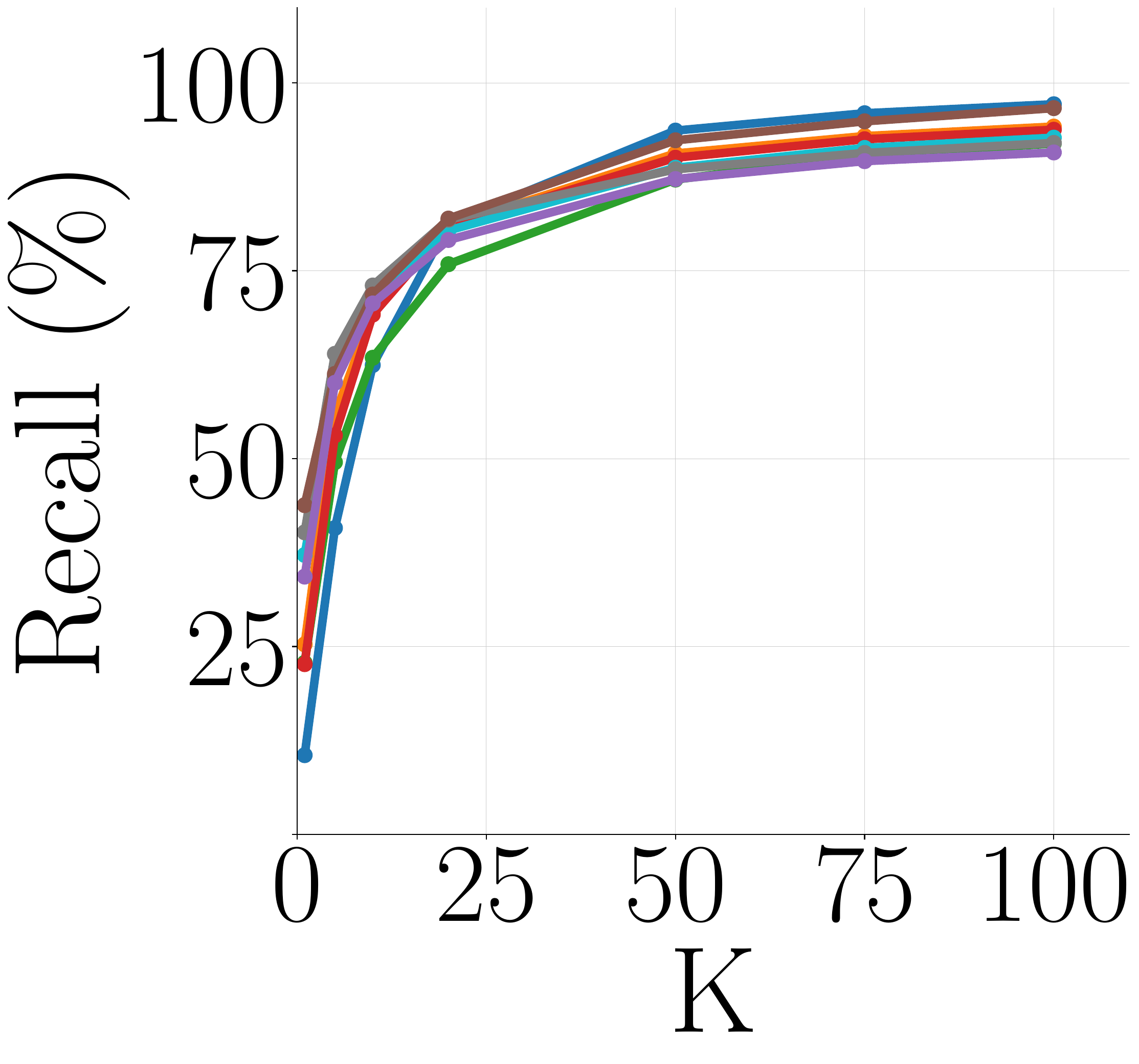} 
&
\includegraphics[width=0.25\linewidth]{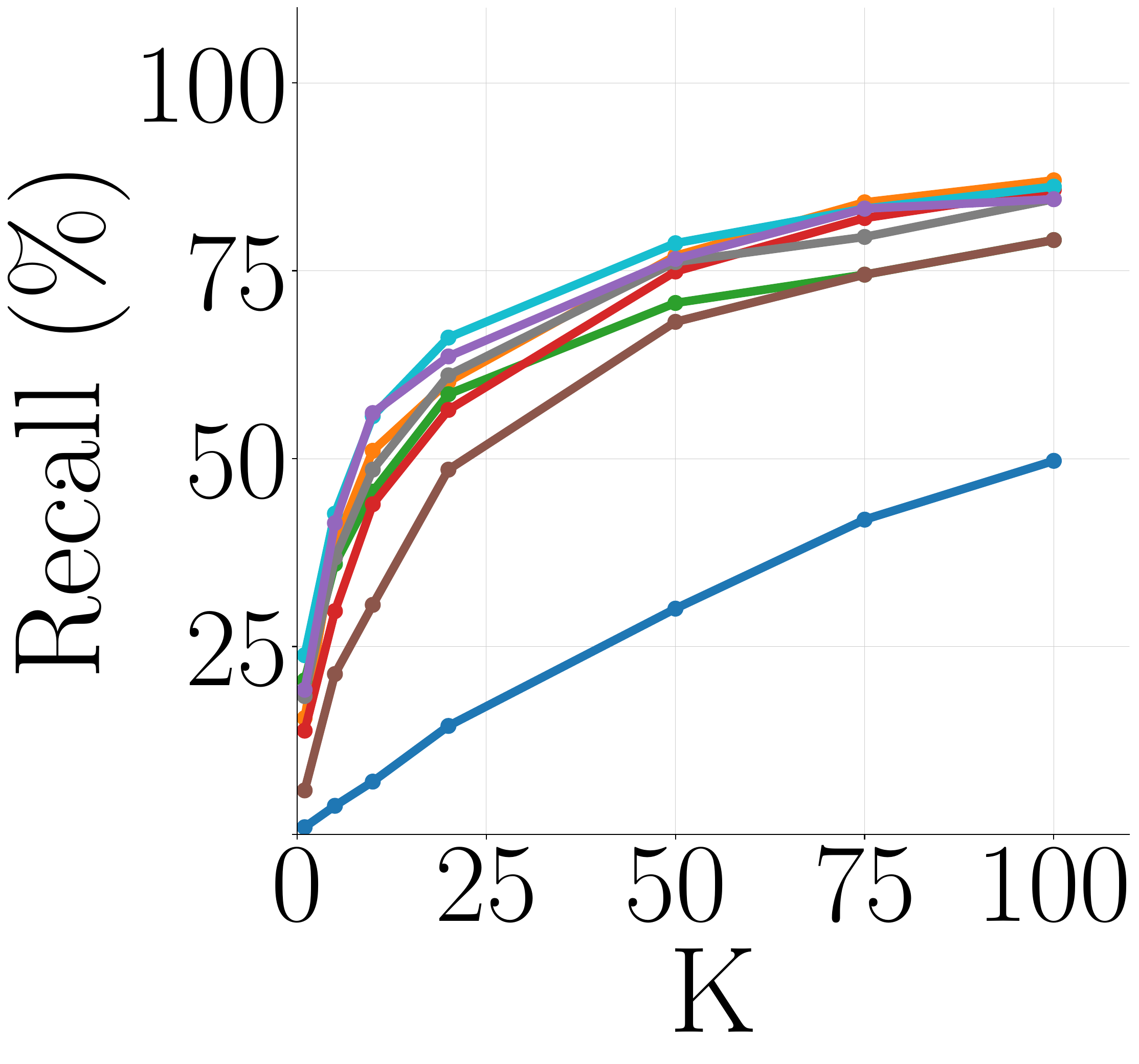} 
&
\includegraphics[width=0.25\linewidth]{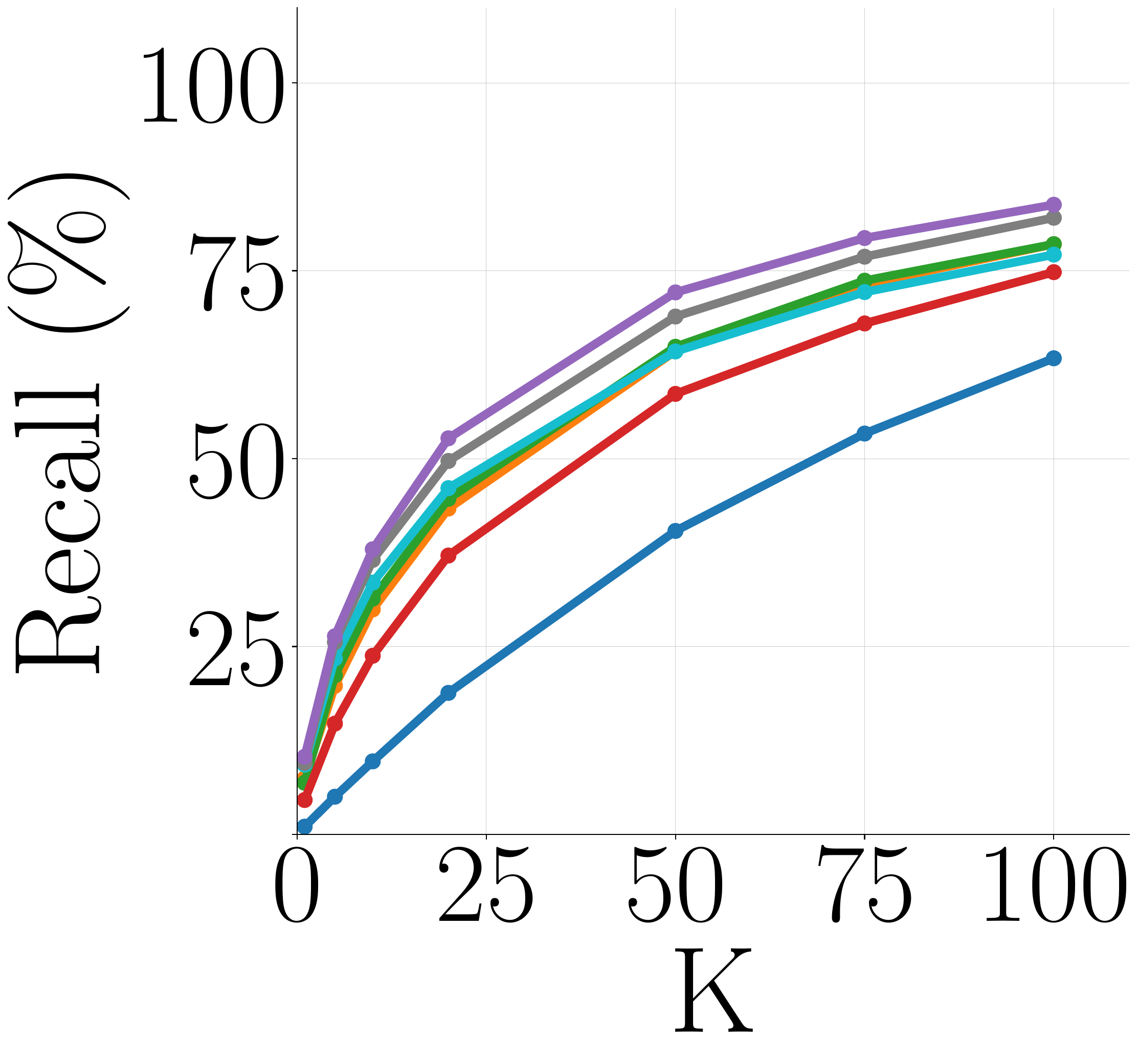} 
\\
(a) Eiffel Tower 2018-2020 & (b) Okinawa 2016-2017 & (c) Okinawa 2017-2018  & (d) Tasman Fracture 2018/12/04-06 & (e) St Helens 2011-2013
\end{tabular}
}
\caption{\textbf{Recall@K performance for VPR methods across underwater datasets.} 
Experimental results illustrating the probability that a correct match appears within the top \( K \) retrieved candidates. We compare six VPR methods—MixVPR, CosPlace, NetVLAD, AnyLoc, MegaLoc, and CricaVPR—against a random guesser to assess whether the retrieval results, given our GPS-based ground truth, correspond to meaningful location identification. Additionally, we include the brute-force SuperPoint approach (which performs feature matching on all possible image pairs)  as an alternative that directly compares local features without global retrieval. In the Okinawa dataset, the random guesser’s performance approaches that of VPR methods for \( K \gtrsim 10 \) due to the lawnmower trajectory pattern (see Fig.~\ref{fig:gps_trajectories}), which increases image density. This allows random selection to occasionally retrieve correct matches, even without visual correspondence. While this does not compromise the GPS-based ground truth, it underscores that VPR methods are most discriminative at lower \( K \) values (see Section~\ref{sub:hierarchical}). Due to its significantly higher computational cost, we do not include the SuperPoint brute-force approach on the St Helens dataset.}
\label{fig:recall_k}
    \vspace*{-0.2cm}
\end{figure*}

\begin{table*}[t]
\centering
\caption{\textbf{Recall@K performance for VPR methods across underwater datasets.} We use green to highlight the {\color{cosplace}best R@10} result for each dataset and orange for the {\color{orange}second-best R@10} result excluding Superpoint Brute-Force. AnyLoc, MegaLoc, and CricaVPR perform comparatively. We chose MegaLoc for subsequent experiments as it is significantly faster than AnyLoc and performs slightly better than CricaVPR on average. We compare results using R@10, as this is the largest value that provides meaningful comparisons given the GPS-based ground truth (see explanation in Figure~\ref{fig:recall_k}).}
\resizebox{\linewidth}{!}{ %
\begin{tabular}{lccccccccccccccc}
\arrayrulecolor{blue10}
& \multicolumn{3}{c}{Eiffel}
& \multicolumn{3}{c}{Okinawa 2016-2017}
& \multicolumn{3}{c}{Okinawa 2017-2018}
& \multicolumn{3}{c}{Tasman Fracture}
& \multicolumn{3}{c}{St Helens} \\ 
\cmidrule(lr){2-4} \cmidrule(lr){5-7} \cmidrule(lr){8-10} \cmidrule(lr){11-13} \cmidrule(lr){14-16}
Methods & R@1 & R@5 & R@10 & R@1 & R@5 & R@10 & R@1 & R@5 & R@10 & R@1 & R@5 & R@10 & R@1 & R@5 & R@10 \\
\cmidrule(lr){1-1} \cmidrule(lr){2-4} \cmidrule(lr){5-7} \cmidrule(lr){8-10} \cmidrule(lr){11-13} \cmidrule(lr){14-16}
MixVPR & 19.2 & 39.7 & 49.5 & 21.9 & 52.6 & 70.9 & 25.3 & 55.8 & 69.5 & 15.5 & 38.5 & 51.1 & 7.5 & 19.8 & 29.9 \\
CosPlace & 15.2 & 32.5 & 42.5 & 27.5 & 51.1 & 64.3 & 22.8 & 49.5 & 63.4 & 20.5 & 35.9 & 45.6 & 6.9 & 21.2 & 31.4 \\
NetVLAD & 5.2 & 17.4 & 25.8 & 25.9 & 56.8 & 73.1 & 22.7 & 53.1 & 69.2 & 13.8 & 29.7 & 43.9 & 4.6 & 14.7 & 23.8 \\
AnyLoc & 18.6 & 38.7 & 48.3 & 46.1 & 73.8 & {\color{cosplace}83.3} & 34.3 & 60.1 & 70.6 & 19.3 & 41.4 & {\color{cosplace}56.0} & 10.3 & 26.4 & {\color{cosplace}37.9} \\ 
MegaLoc & 33.7 & 55.1 & {\color{cosplace}65.1} & 47.2 & 72.0 & {\color{orange}79.7} & 40.2 & 63.9 & {\color{cosplace}73.0} & 18.4 & 36.8 & 48.5 & 9.5 & 25.6 & {\color{orange}36.5} \\
CricaVPR & 25.1 & 47.7 & {\color{orange}58.1} & 32.6 & 59.5 & 73.1 & 37.2 & 60.2 &{\color{orange}71.0} & 23.8 & 42.7 & {\color{orange}55.6} & 9.3 & 23.4 & 33.5 \\
\multicolumn{16}{c}{\color{blue10}- - - - - - - - - - - - - - - - - - - - - - - - - - - - - - - - - - - - - - - - - - - - - - - - - - - - - - - - - - - - - - - - - - - - - - - - - - - - - - - - - - - - - - - - - - - - - - - - -} \\
Superpoint Brute-Force & 28.9 & 54.3 & 63.1 & 19.6 & 51.5 & 71.3 & 43.8 & 61.3 & 71.8  & 5.9 & 21.3 & 30.5 & - & - & -  \\ 
\hline
\end{tabular}
}
\label{tab:vpr_recall}
    \vspace*{-0.2cm}
\end{table*}

\begin{table}[t]
    \centering
    \caption{\textbf{Average computation time per query (ms) for each VPR method.} All one-stage VPR techniques are significantly faster than the brute-force SuperPoint approach. Our hierarchical method, combining MegaLoc and SuperPoint, is \( 100\times \) faster than brute-force SuperPoint.}
    \begin{tabular}{lc}
    \arrayrulecolor{blue10}
        Methods & Avg. Time per Query (ms) \\
        \cmidrule(lr){1-1} \cmidrule(lr){2-2}
        MixVPR & 8 \\
        CosPlace & 14 \\
        NetVLAD & 21 \\
        AnyLoc & 428 \\ 
        MegaLoc & 49 \\
        CricaVPR & 38 \\
        Superpoint-B & 27150 \\ 
        \multicolumn{2}{c}{\color{blue10}- - - - - - - - - - - - - - - - - - - - - - - - - - - - - - - - - -} \\ 
        MegaLoc + Superpoint-H & 49 + 226 \\
        \hline
    \end{tabular}
    \label{tab:times}
    \vspace*{-0.3cm}
\end{table}

\subsection{VPR Baselines}
\label{sec:vprBaselines}

For global retrieval, we evaluate state-of-the-art visual place recognition (VPR) methods implemented in~\cite{Berton_2023_EigenPlaces}, including MixVPR~\cite{ali2023mixvpr}, CosPlace~\cite{Berton:Cosplace:2022}, NetVLAD~\cite{Arandjelović:NetVLAD:2018}, AnyLoc~\cite{Keetha:AnyLoc:2024}, MegaLoc~\cite{berton2025megalocretrievalplace}, and CricaVPR~\cite{lu2024cricavpr}. While our pipeline supports all models referenced in~\cite{Berton_2023_EigenPlaces}, we select this subset due to their strong performance and widespread adoption in terrestrial VPR. For local feature refinement, we use SuperPoint~\cite{detone2018superpoint} keypoints in combination with LightGlue~\cite{Lindenberger:LightGlue:2023}.  

To assess whether the retrieval results, based on GPS-derived ground truth, correspond to meaningful place recognition, we compare the evaluated VPR methods against a random guesser, which serves as a lower performance bound. The random guesser is implemented as a Monte Carlo experiment: for each query image \( Q_i \), we randomly select \( K \) database matches, repeating this selection \( n \) times per query. The Recall@K metric is then defined as the proportion of iterations where at least one correct match was retrieved among the \( K \) selected database images, normalized by the total number of trials, $|Q| \cdot  n $, where $|Q|$ is the number of query images and \( n \) is chosen to be sufficiently large. We also compare to a brute-force baseline that performs exhaustive local feature matching on every query-database image pair using LightGlue. In this case, the number of keypoint matches serves as the similarity measure between image pairs.

\subsection{SQUIDLE+ VPR Benchmark}

In this work, we propose using data from SQUIDLE+ to advance underwater VPR by leveraging a vast collection of unstructured data. This dataset originates from various robotic platforms executing diverse trajectory patterns with arbitrary overlap, spanning time differences ranging from days to years. The data includes diverse seafloor types captured under varying out-of-distribution scene conditions (e.g., depth, lighting, turbidity). Table~\ref{tab:datasets} provides details of the selected sequences, while Figure~\ref{fig:gps_trajectories} illustrates the robot trajectories and presents sample RGB images depicting the scene variations.

We process the sequences by selecting only those captured with downward-looking cameras, the most common setup for exploring seafloor environments. This distinguishes our benchmark from most datasets in structured environments captured with forward-facing cameras, where the majority of VPR techniques are trained~\cite{Keetha:AnyLoc:2024}, making it more comparable to aerial datasets~\cite{schleiss2022vpair}.

To ensure consistent processing times across datasets, we downsample the RGB images to a resolution of approximately 640$\times$480 pixels, adjusted to maintain the original aspect ratio. We include AUV sequences with GPS signals to evaluate our retrieval pipeline using an image-independent ground truth for visual localization assessment~\cite{brachmann2021limits}. 

\noindent  Our dataset includes the following sequences:

\textbf{Okinawa}\footnote{\scriptsize \url{https://oceanperception.com/impact/expeditions/}}: This dataset captures the mesophotic coral seafloor off Sesoko Island, near Okinawa, over a two-year period using the AUV \textit{Tuna-Sand}~\cite{nakatani2008auv}. The 2018 sequence was recorded days after impact from Typhoon Trami (Paeng).  Visual changes from environmental damage hinder relocalization, emphasizing the need for a robust pipeline to document damage and recovery.

\textbf{Tasman Fracture}\footnote{\scriptsize \url{https://data.csiro.au/collection/60886}}:
This dataset consists of benthic stereo still imagery collected to map the distribution of epibenthic fauna and habitats on seamounts off Tasmania~\cite{untiedt2023tasmanian}. The imagery was acquired at a depth of 885 meters over a two-day period using a deep-towed camera.

\textbf{St Helens}\footnote{\scriptsize \url{https://www.researchgate.net/publication/294422138_Stereo-imaging_AUV_detects_trends_in_sea_urchin_abundance_on_deep_overgrazed_reefs}}: This dataset comprises benthic stereo-imaging surveys conducted with the AUV Sirius to monitor long-term urchin population dynamics in St Helens, northeast Tasmania~\cite{ling2016stereo}. As part of a bi-yearly monitoring program, the AUV surveyed shallow barren/kelp transition zones, which are included in our benchmark.

\begin{figure*}[t]
\centering
\resizebox{\linewidth}{!}{ %
\begin{tabular}{ccccc}
\multicolumn{5}{c}{\myline{mixvpr} MIXVPR {\myline{cosplace}} COSPLACE {\myline{netvlad}} NETVLAD {\myline{anyloc}} ANYLOC {\myline{megaloc}} MEGALOC {\myline{crica}} CRICAVPR {\myline{superpointbrute}} SUPERPOINT-B {\myline{hierarchical}} SUPERPOINT-H} 
\\
\includegraphics[width=0.22\linewidth]{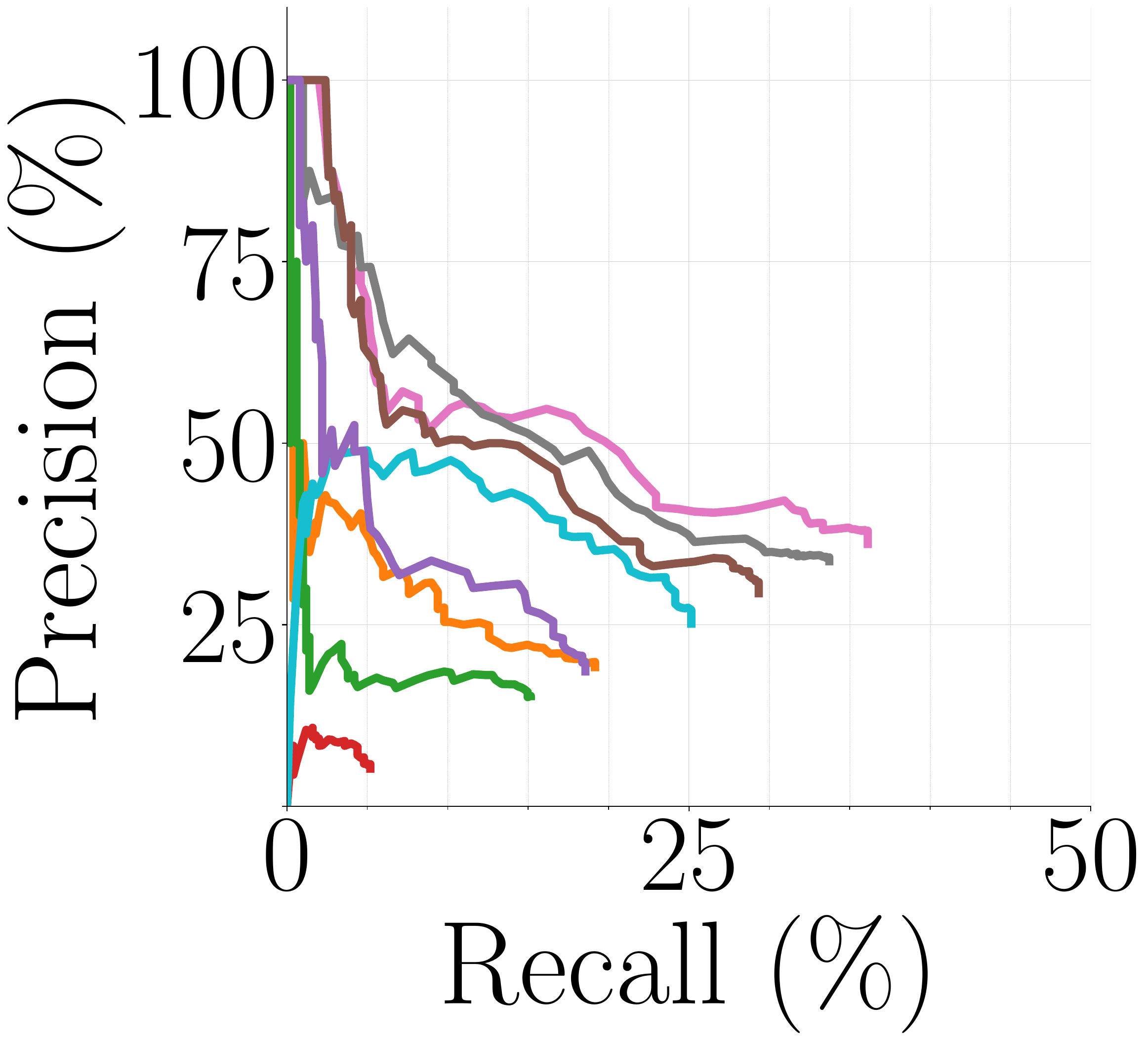} 
&
\includegraphics[width=0.22\linewidth]{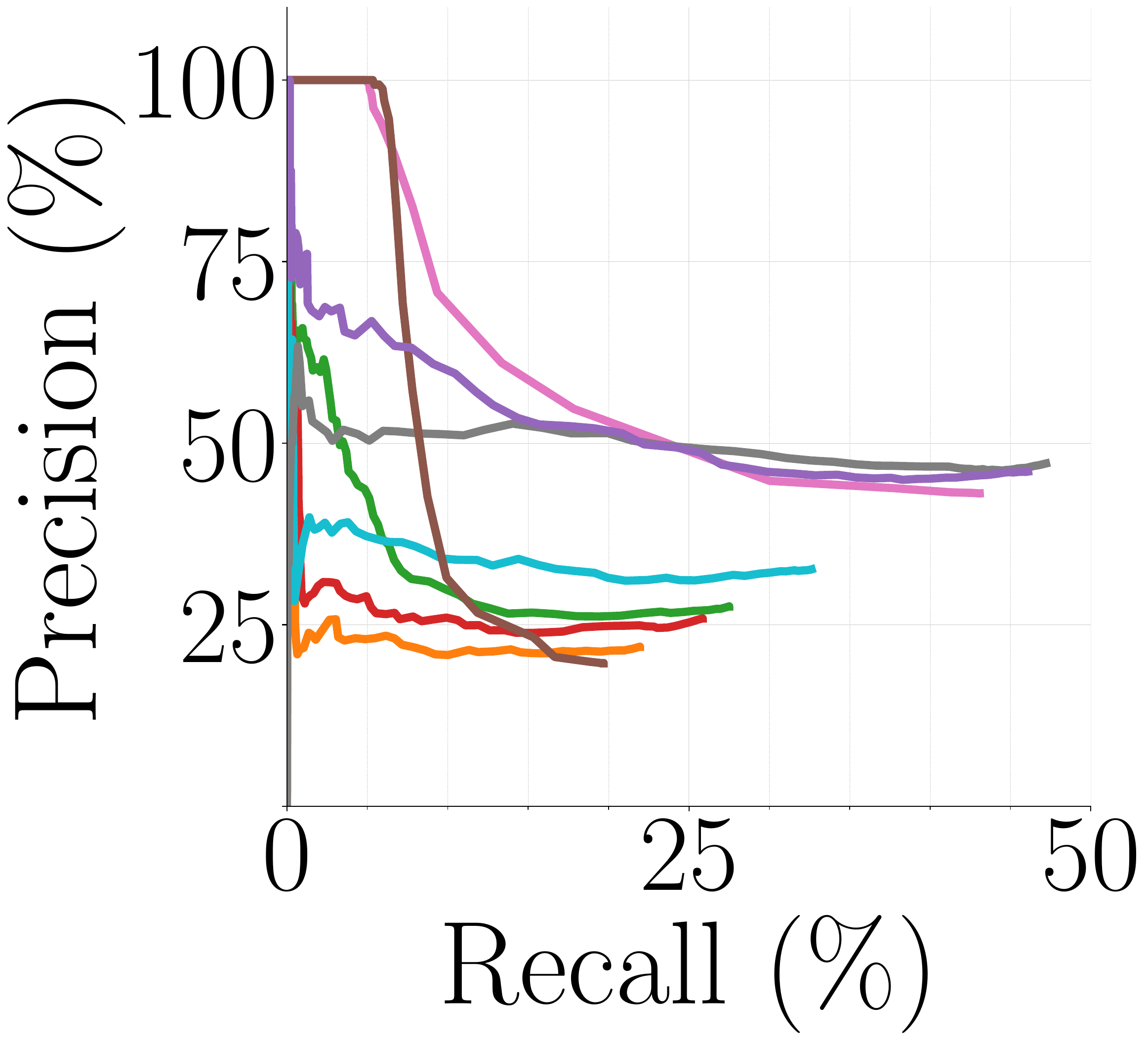} 
&
\includegraphics[width=0.22\linewidth]{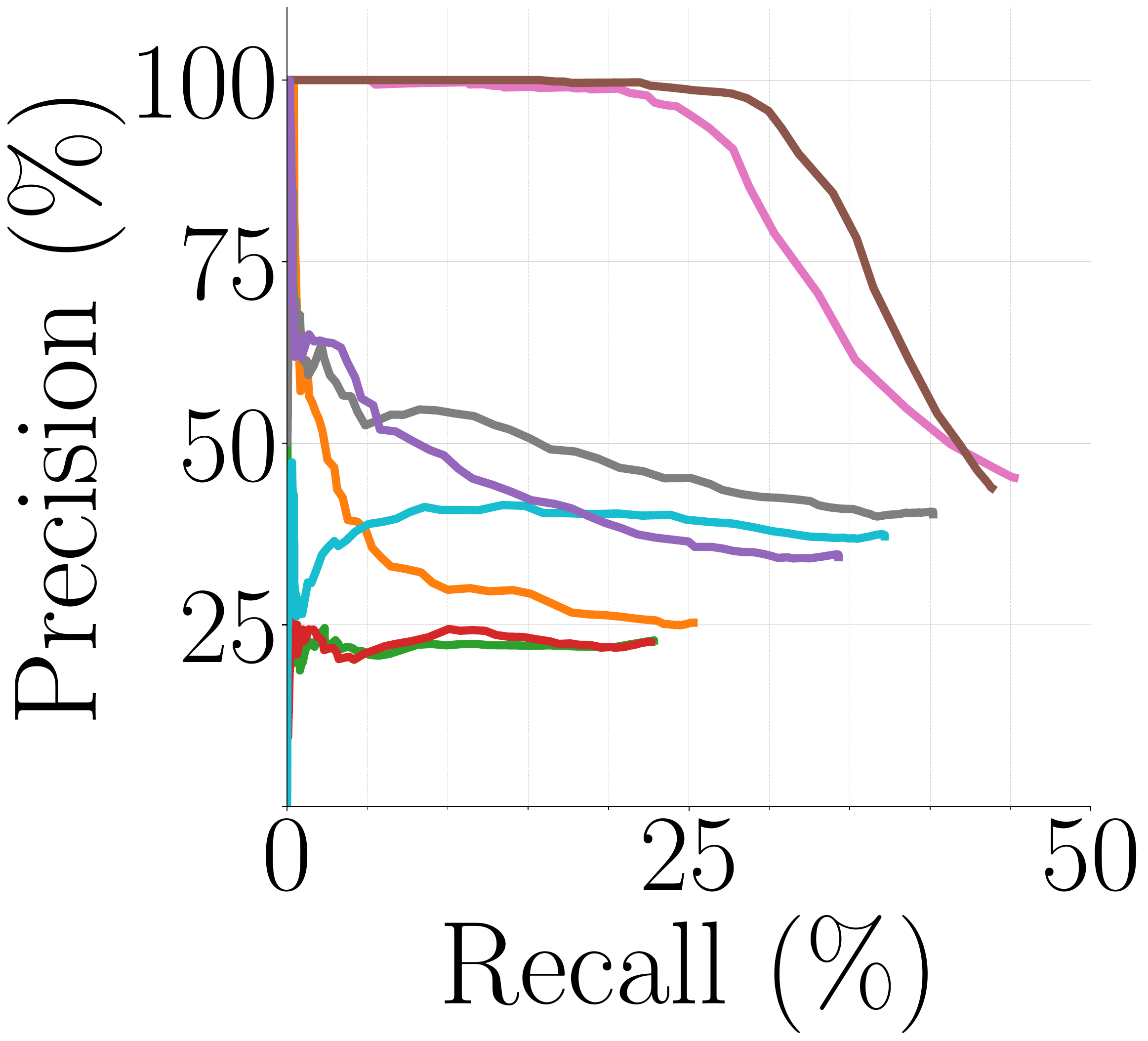} 
&
\includegraphics[width=0.22\linewidth]{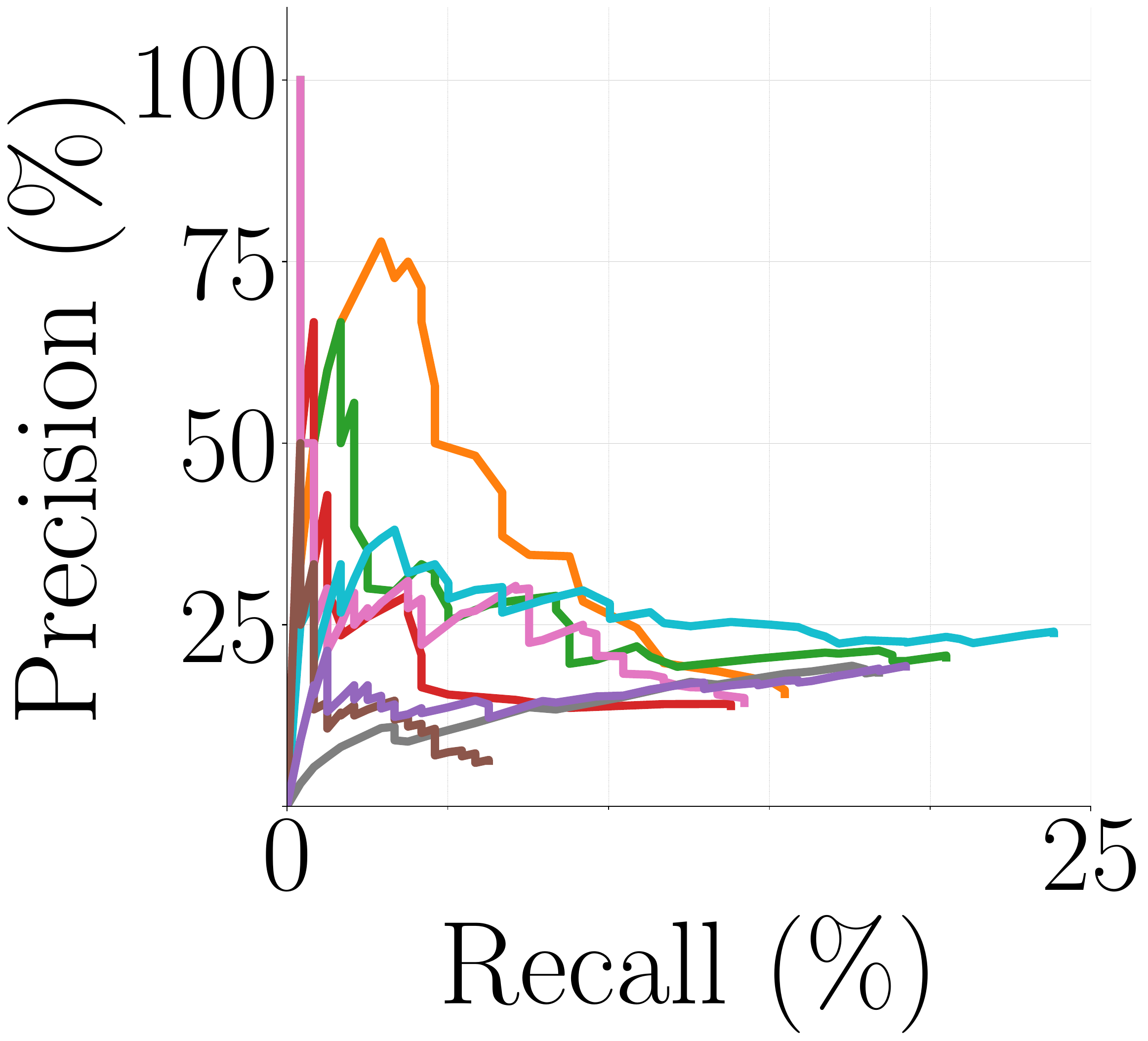} 
&
\includegraphics[width=0.22\linewidth]{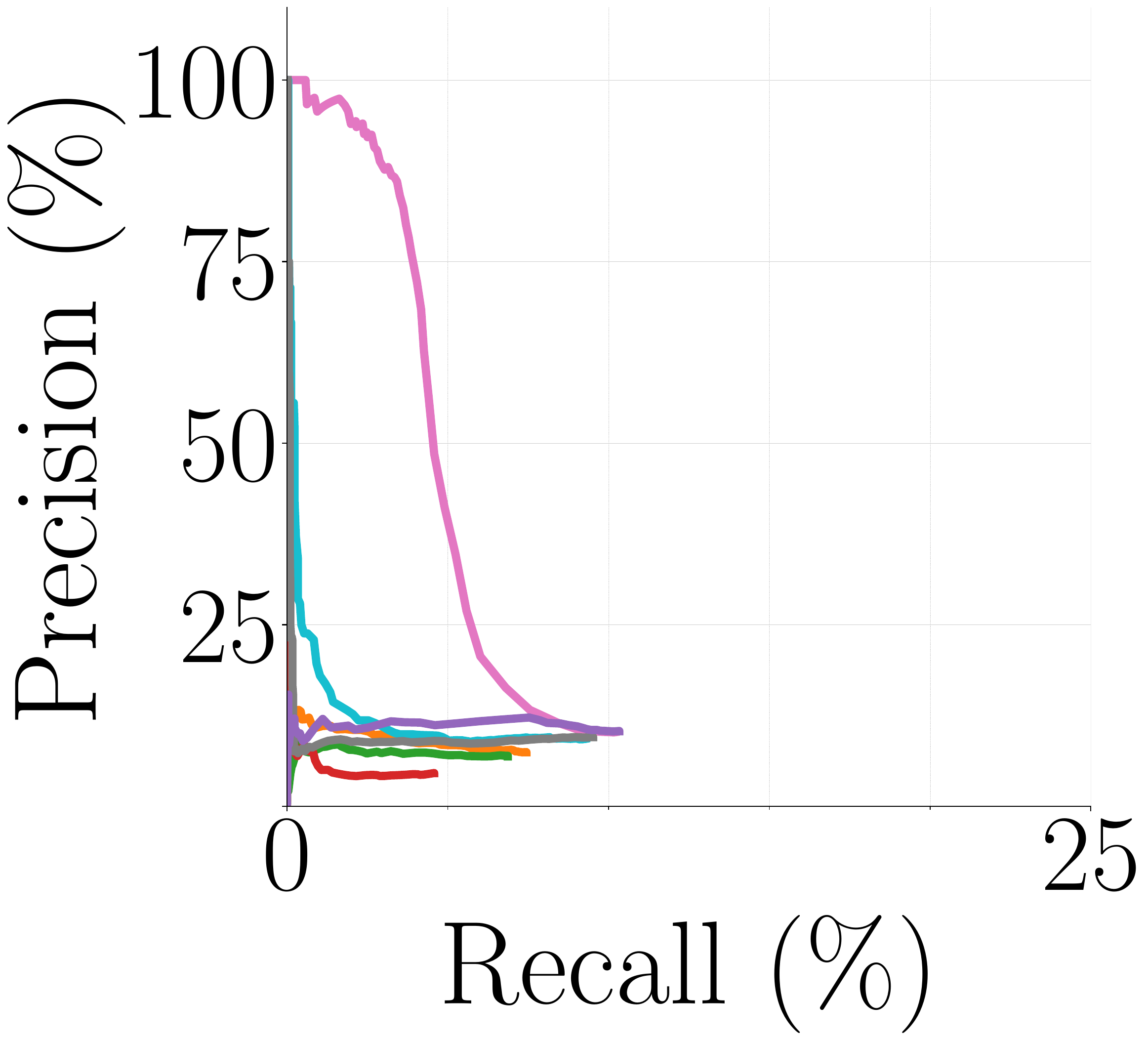} 
\\
\includegraphics[width=0.22\linewidth, height=3cm]{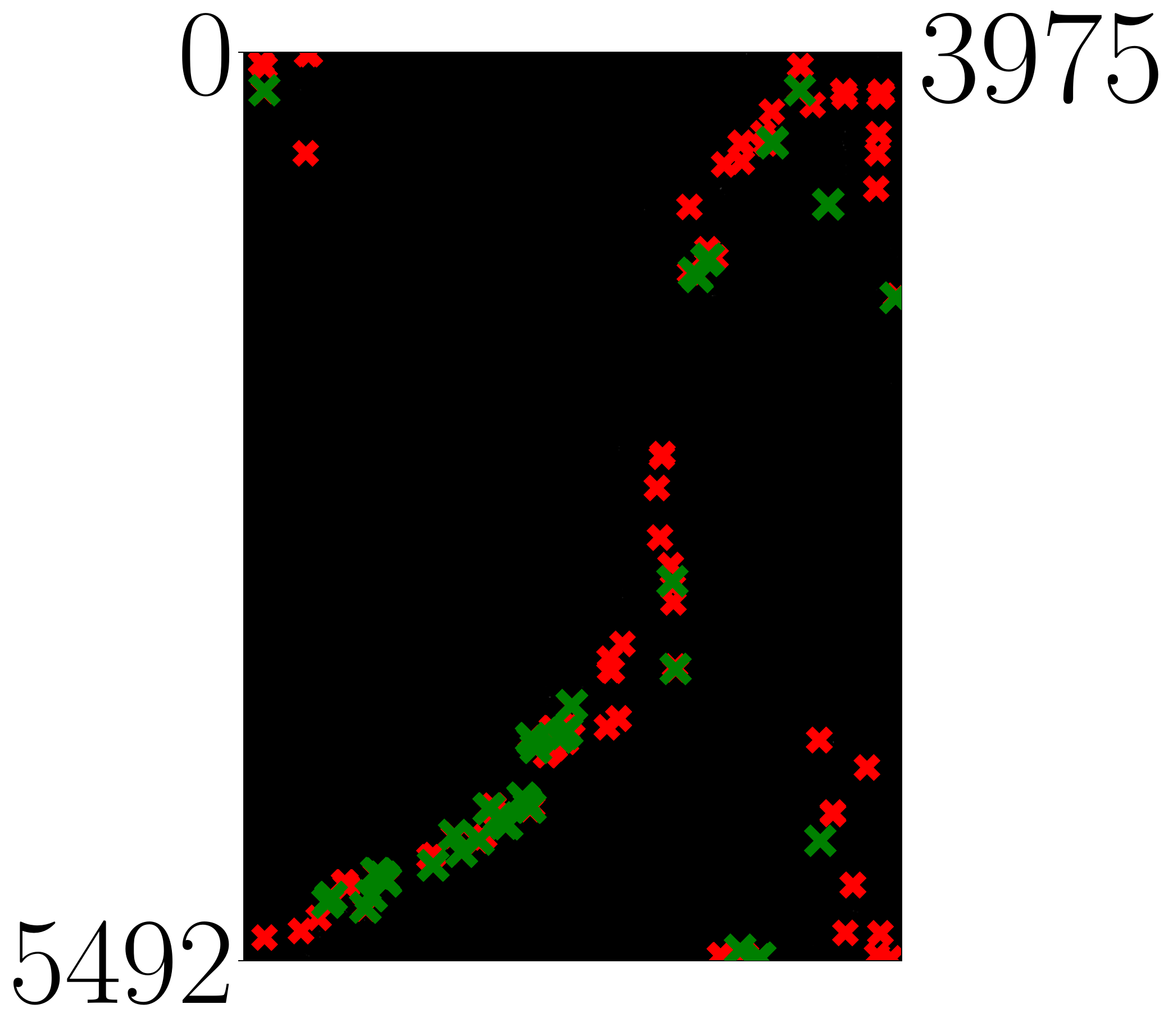}
&
\includegraphics[width=0.22\linewidth, height=3cm]{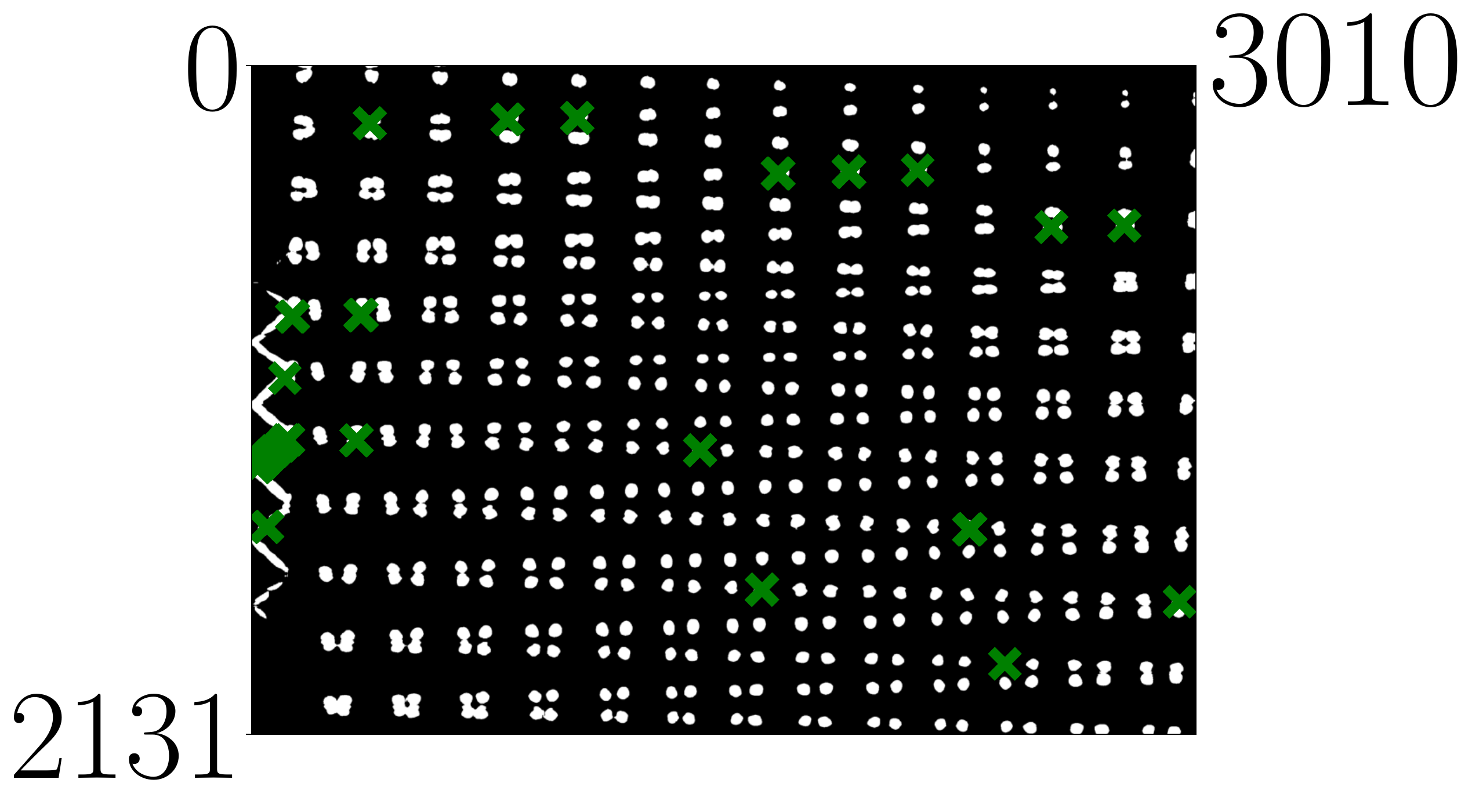} 
&
\includegraphics[width=0.22\linewidth, height=3cm]{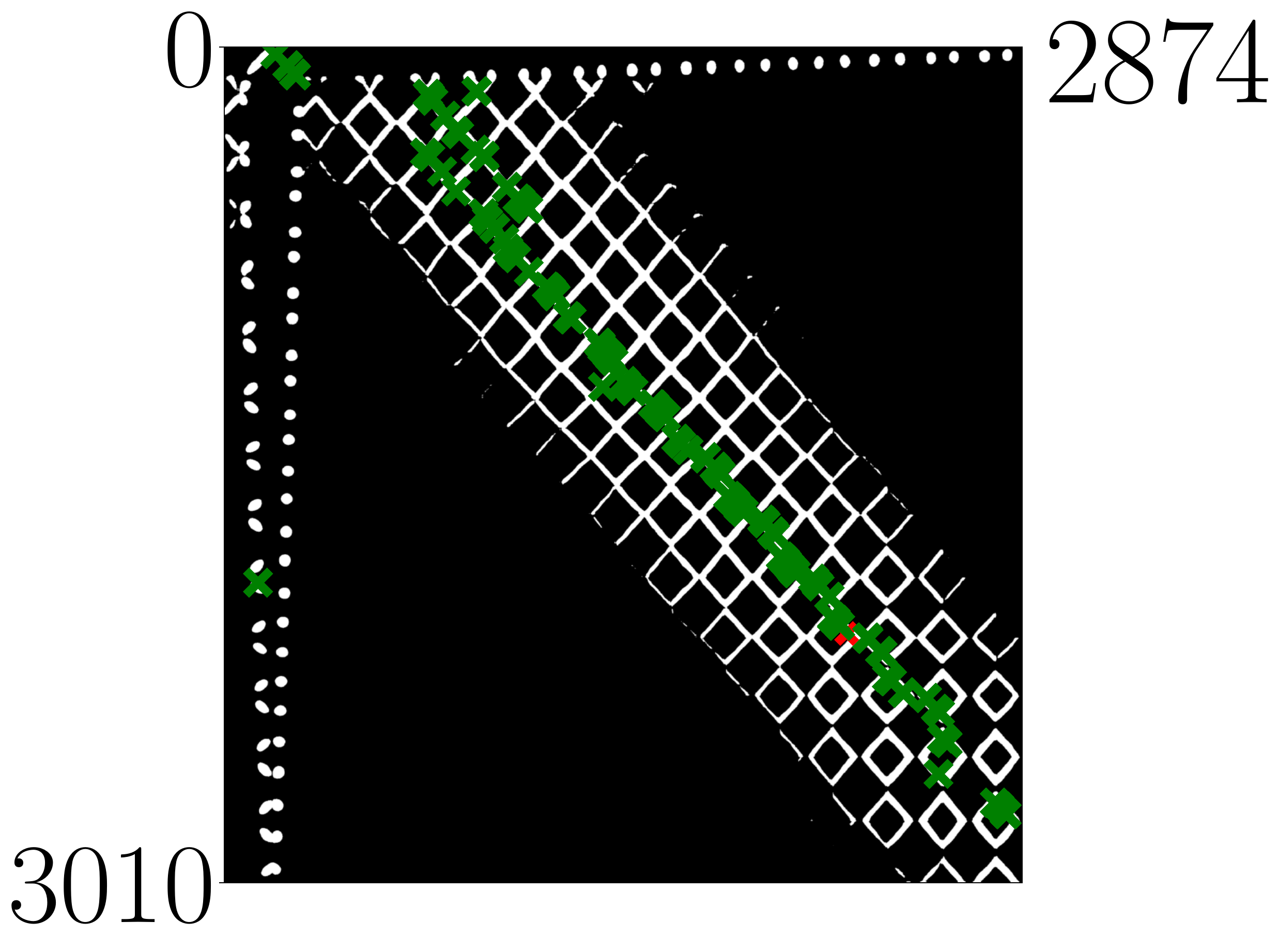}
&
\includegraphics[width=0.22\linewidth, height=3cm]{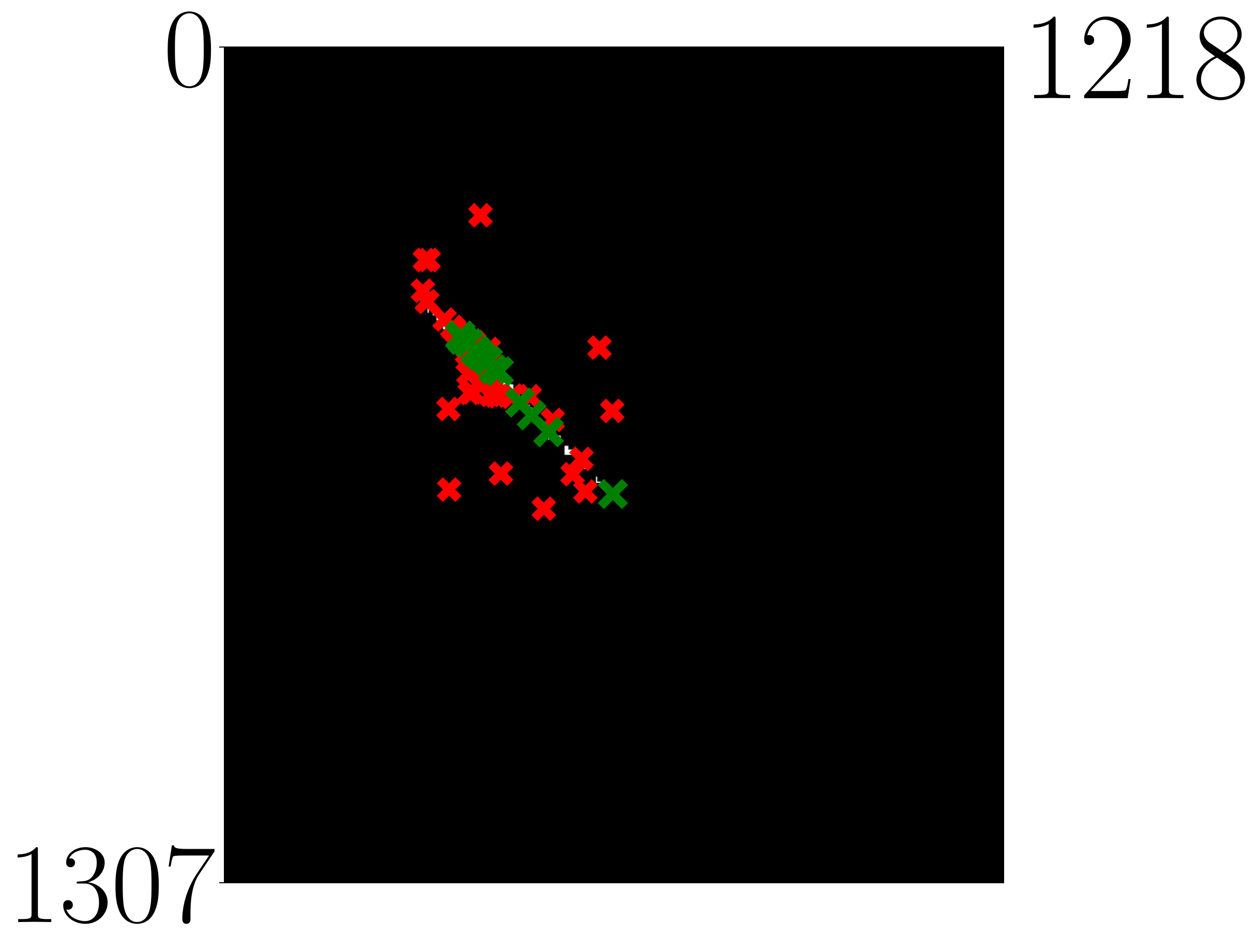} 
&
\includegraphics[width=0.22\linewidth, height=3cm]{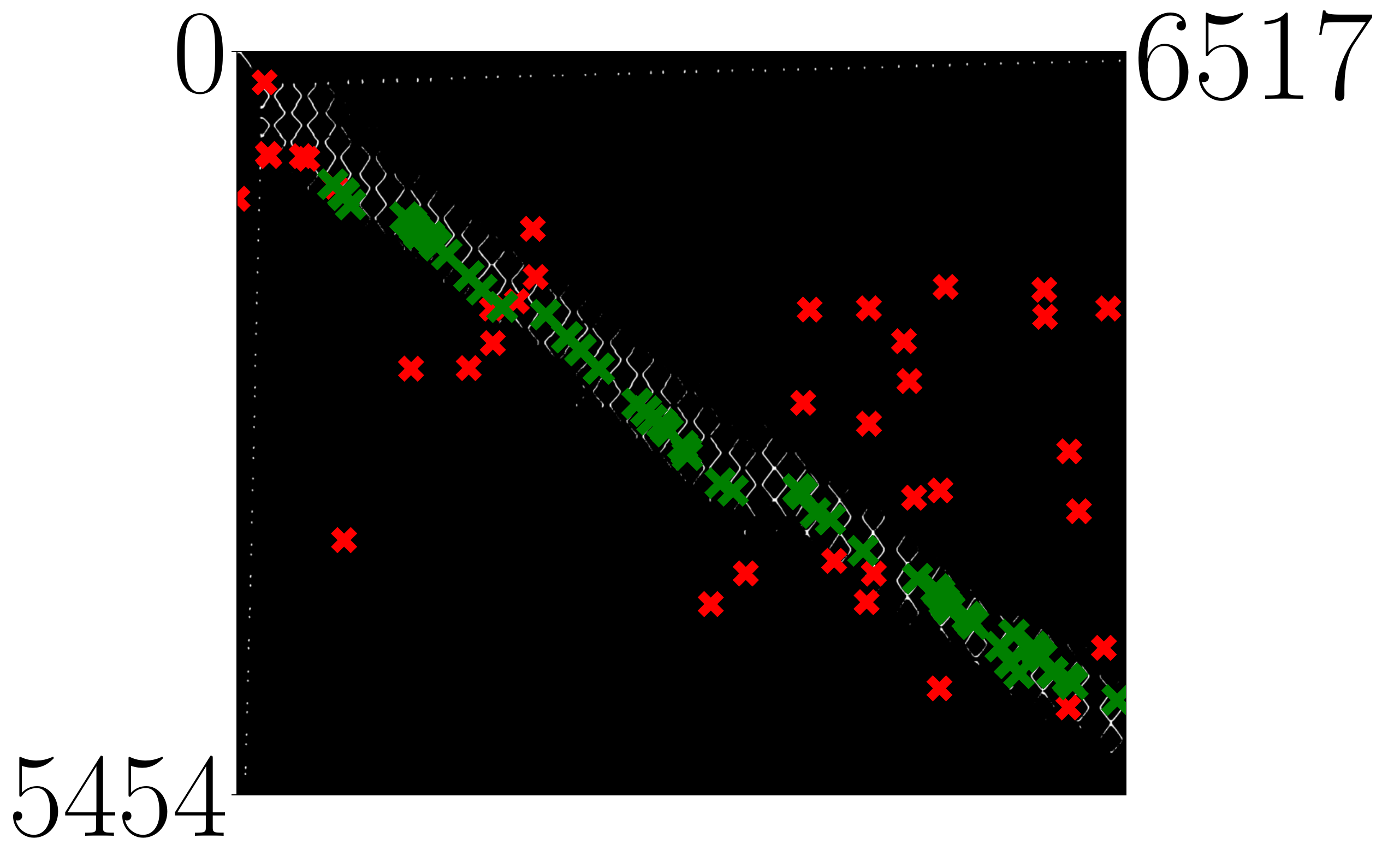} 
\\
(a) Eiffel Tower 2018-2020 & (b) Okinawa 2016-2017 & (c) Okinawa 2017-2018 & (d) Tasman Fracture 2018/12/04-06 & (e) St Helens 2011-2013
\end{tabular}
}
\caption{\textbf{Top row:} Precision-Recall curves for Best-Single-Match VPR. Our hierarchical {\color{hierarchical}\textit{SuperPoint-H}} method outperforms one-stage VPR approaches, achieving performance closer to the {\color{superpointbrute}{SuperPoint brute-force}} approach while significantly reducing computational cost (see Table~\ref{tab:times}). \textbf{Bottom row:} The background represents the binary ground truth in black and white, with green crosses indicating {\color{Green}true positives} and red crosses indicating {\color{red}false positives} for the best match per query using our {\color{hierarchical}\textit{SuperPoint-H}} method. Positive matches are filtered to exclude those with a reprojection error greater than 10 pixels, corresponding to a precision of 39\% for the Eiffel Tower dataset, 99\% for Okinawa, 22\% for Tasman Fracture, and 72\% for St Helens.}

\label{fig:pr_curves}
    \vspace*{-0.3cm}
\end{figure*}

We include the \textbf{Eiffel Tower}\footnote{\scriptsize \url{https://www.seanoe.org/data/00810/92226/}}~\cite{Boittiaux:EiffelTower:2023} dataset alongside the SQUIDLE+ sequences to provide a curated reference specifically designed for benchmarking long-term underwater visual localization. The dataset consists of images captured during four visits to the same hydrothermal vent edifice over a five-year period. Camera poses and a common scene geometry were estimated using navigation data and SfM. In contrast to~\cite{Keetha:AnyLoc:2024}, where only ${\sim}1\%$ of the images were used, we incorporate entire sequences for a more comprehensive evaluation. 

For all sequences, we obtain the ground truth matrix where each pair of images is considered to depict the same place if their spatial distance is smaller than a predefined localization radius (see Table~\ref{tab:datasets}).

\subsection{Evaluation Metrics}
We assess VPR methods using established metrics, as they are widely recognized in the field and provide a clear framework for evaluating performance. This approach enables us to contextualize the behavior of the methods within the standards of the research community while also highlighting their effectiveness on challenging underwater datasets. We do not analyze the impact of image resolution on VPR performance, as it falls outside the scope of this work (see~\cite{tomita2023visualplacerecognitionlowresolution} for a discussion).

\textbf{Recall@K}: The probability that at least one of the top K matches, ranked by similarity for a given query, is a correct match (i.e., a true positive).

\textbf{Precision-Recall Curves:} A precision-recall curve evaluates VPR performance when considering only the highest-ranked match per query. This analysis highlights the trade-off between precision and recall, providing insights into the reliability of the top-ranked match in correctly identifying revisited locations.

\section{Results and Discussion}

\begin{figure}[t]
\centering
\resizebox{\linewidth}{!}{ %
\begin{tabular}{cccccc}
\\ 
&  \LARGE Query & \LARGE Database & \LARGE Warping & \LARGE IoU
\\
\multirow{2}{*}{\rotatebox{90}{\LARGE Okinawa \hspace{-0.75cm}}}
&\includegraphics[width=0.50\linewidth]{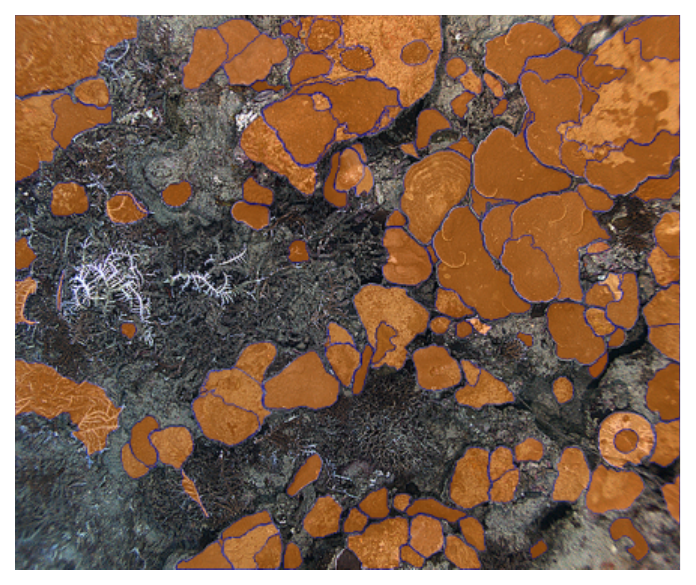}
&\includegraphics[width=0.50\linewidth]{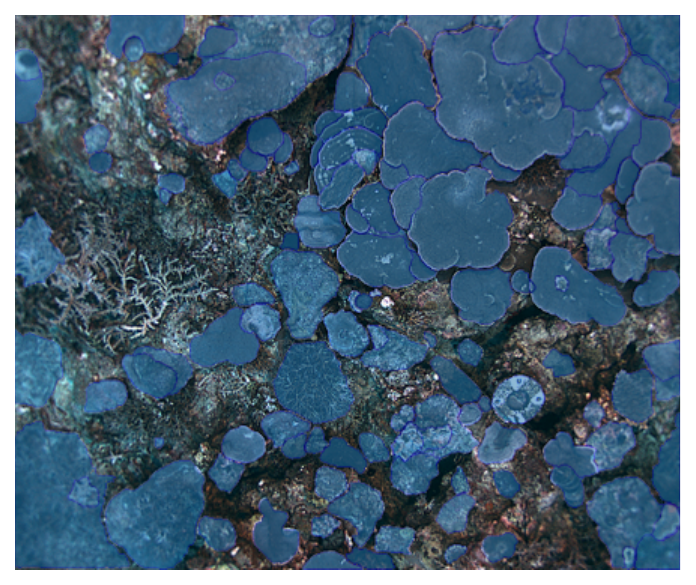} 
&\includegraphics[width=0.50\linewidth]{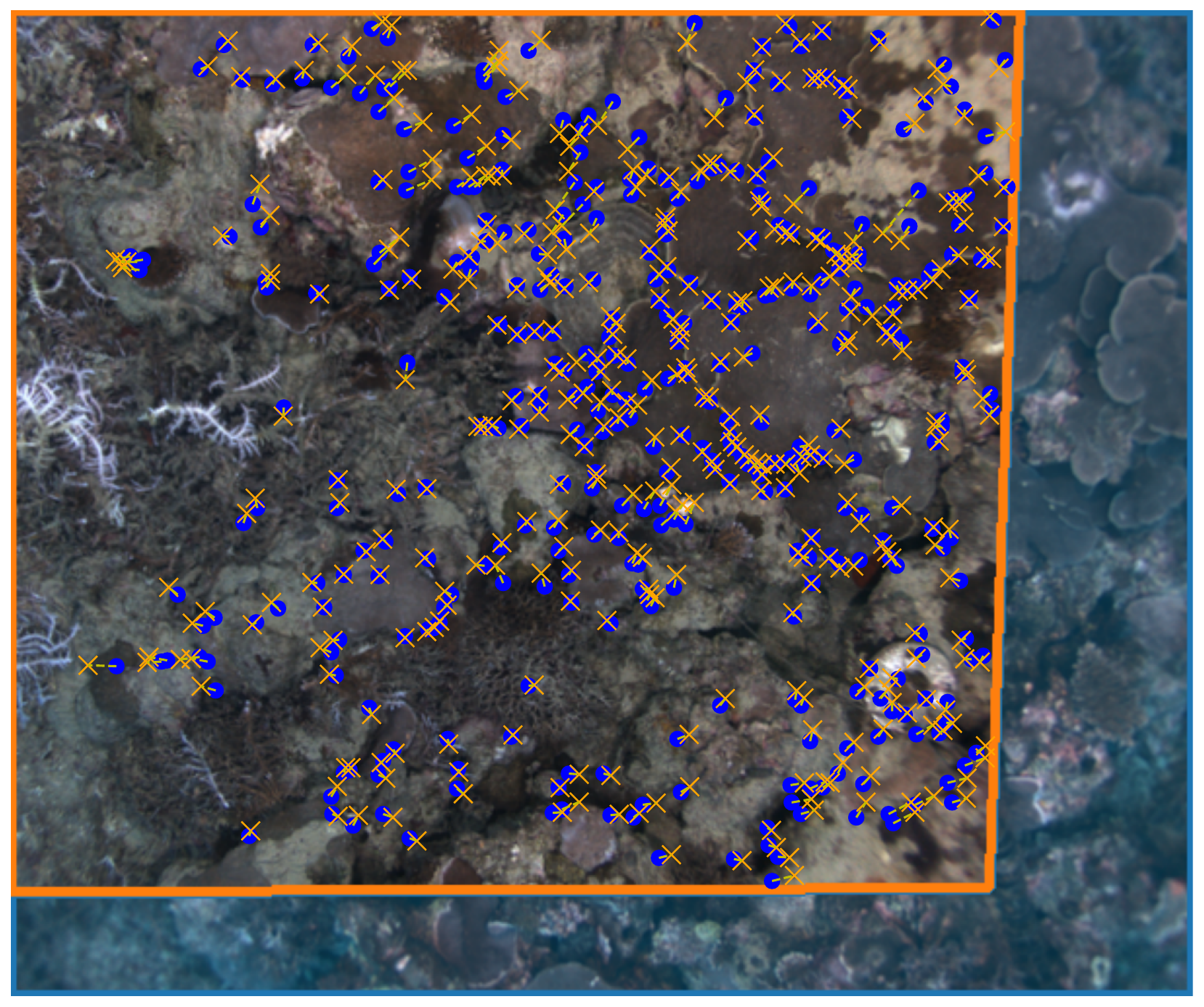} 
&\includegraphics[width=0.50\linewidth]{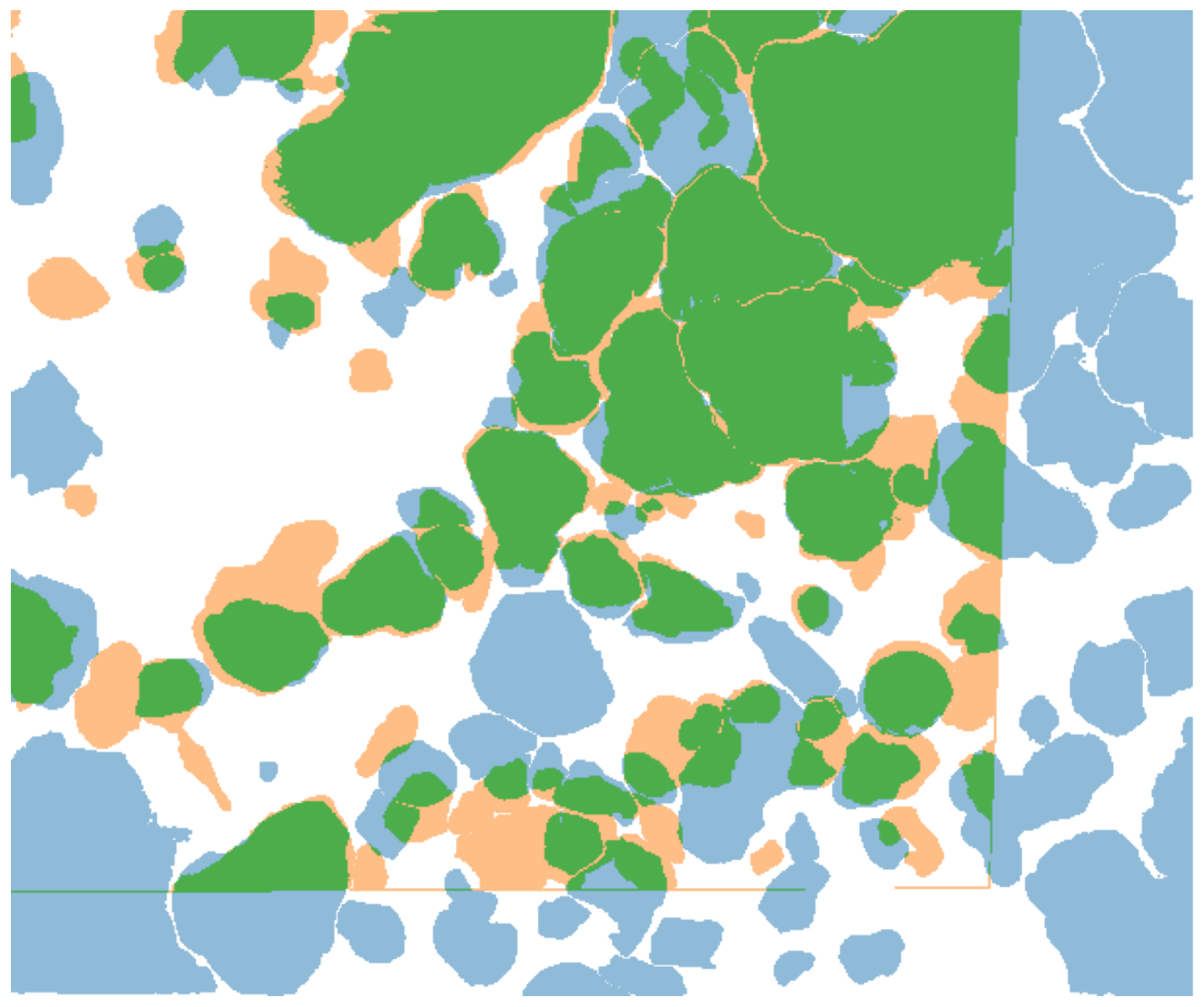}
& {\rotatebox{90}{\LARGE \hspace{1cm}0.65}}
\\
&\includegraphics[width=0.50\linewidth]{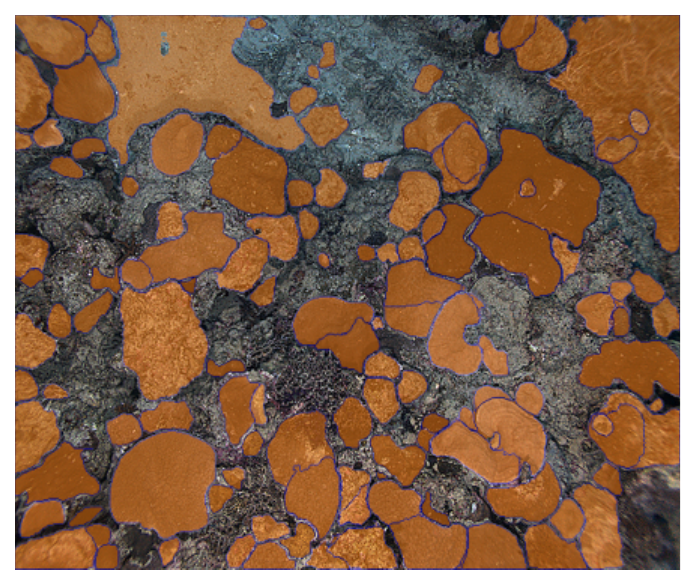}
&\includegraphics[width=0.50\linewidth]{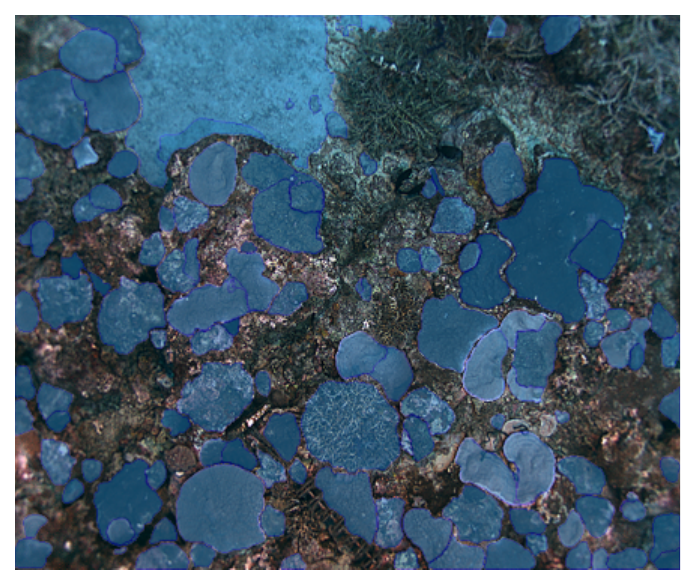} 
&\includegraphics[width=0.50\linewidth]{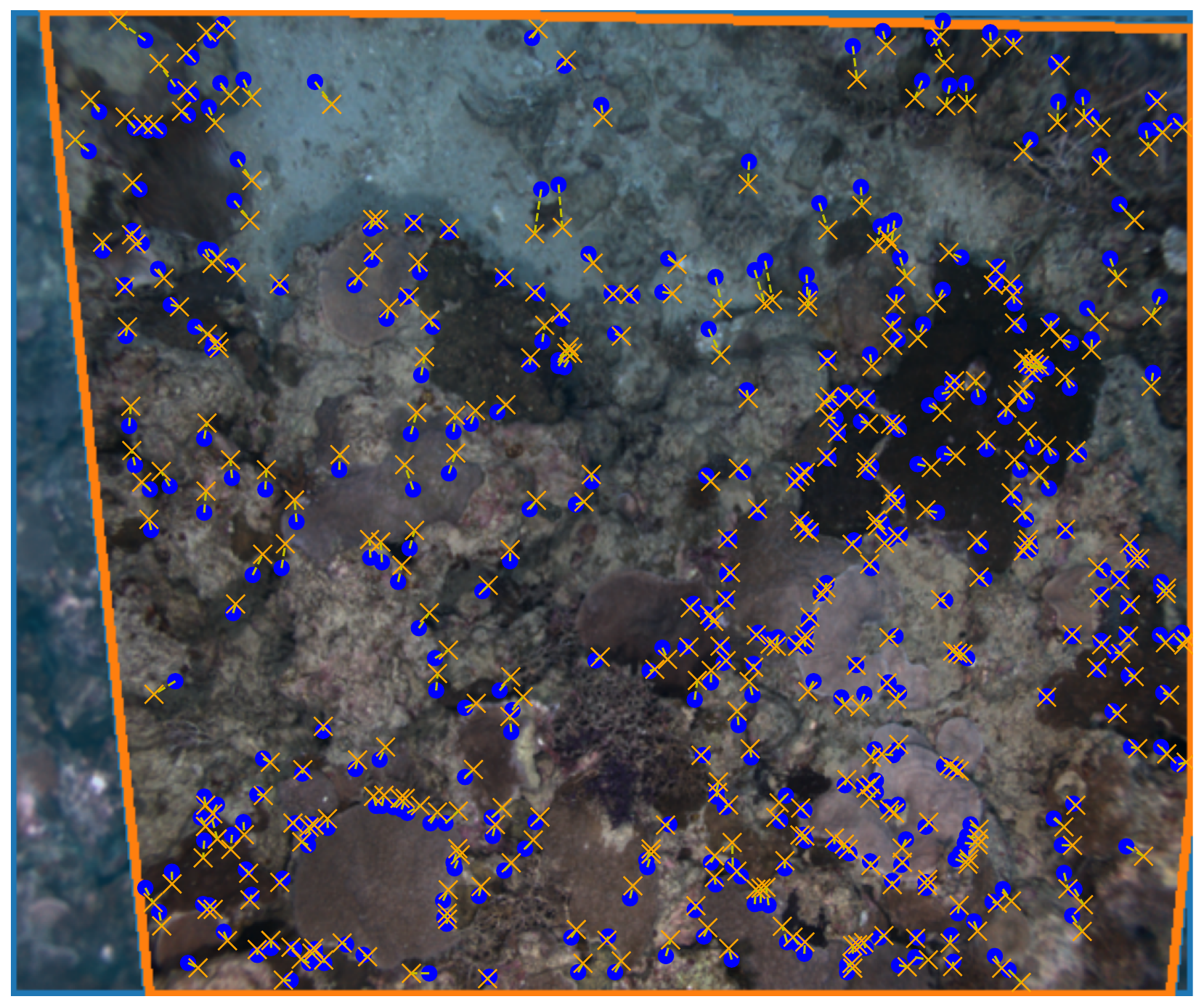} 
&\includegraphics[width=0.50\linewidth]{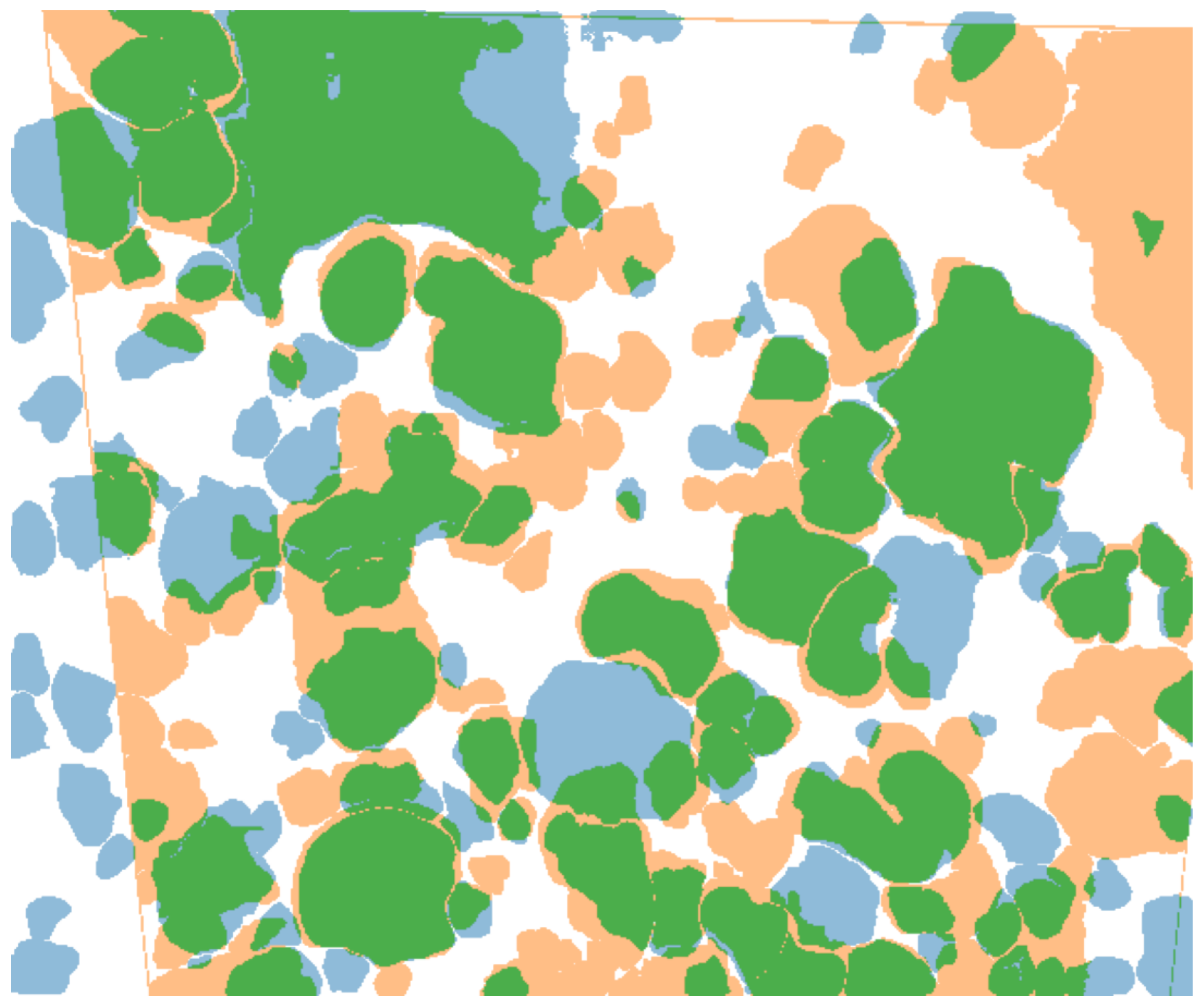}
& {\rotatebox{90}{\LARGE \hspace{1cm}0.65}}
\\
\multirow{2}{*}{\rotatebox{90}{\LARGE Tasman Fracture \hspace{-1.75cm}}}
&\includegraphics[width=0.50\linewidth]{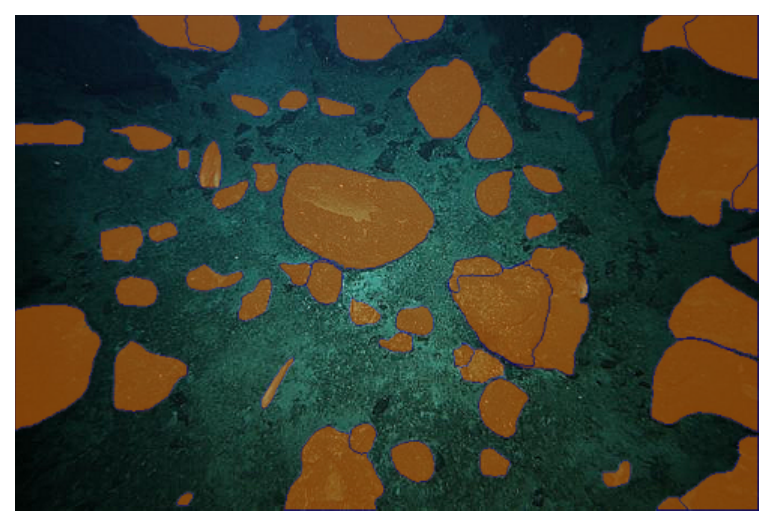}
&\includegraphics[width=0.50\linewidth]{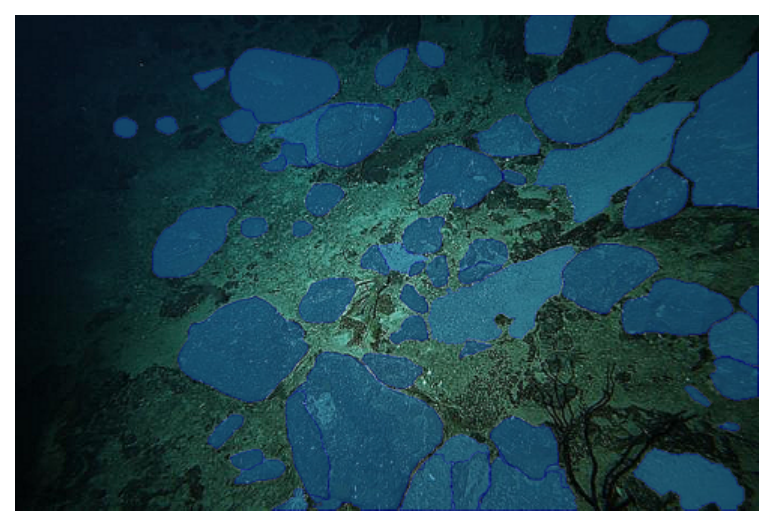} 
&\includegraphics[width=0.50\linewidth]{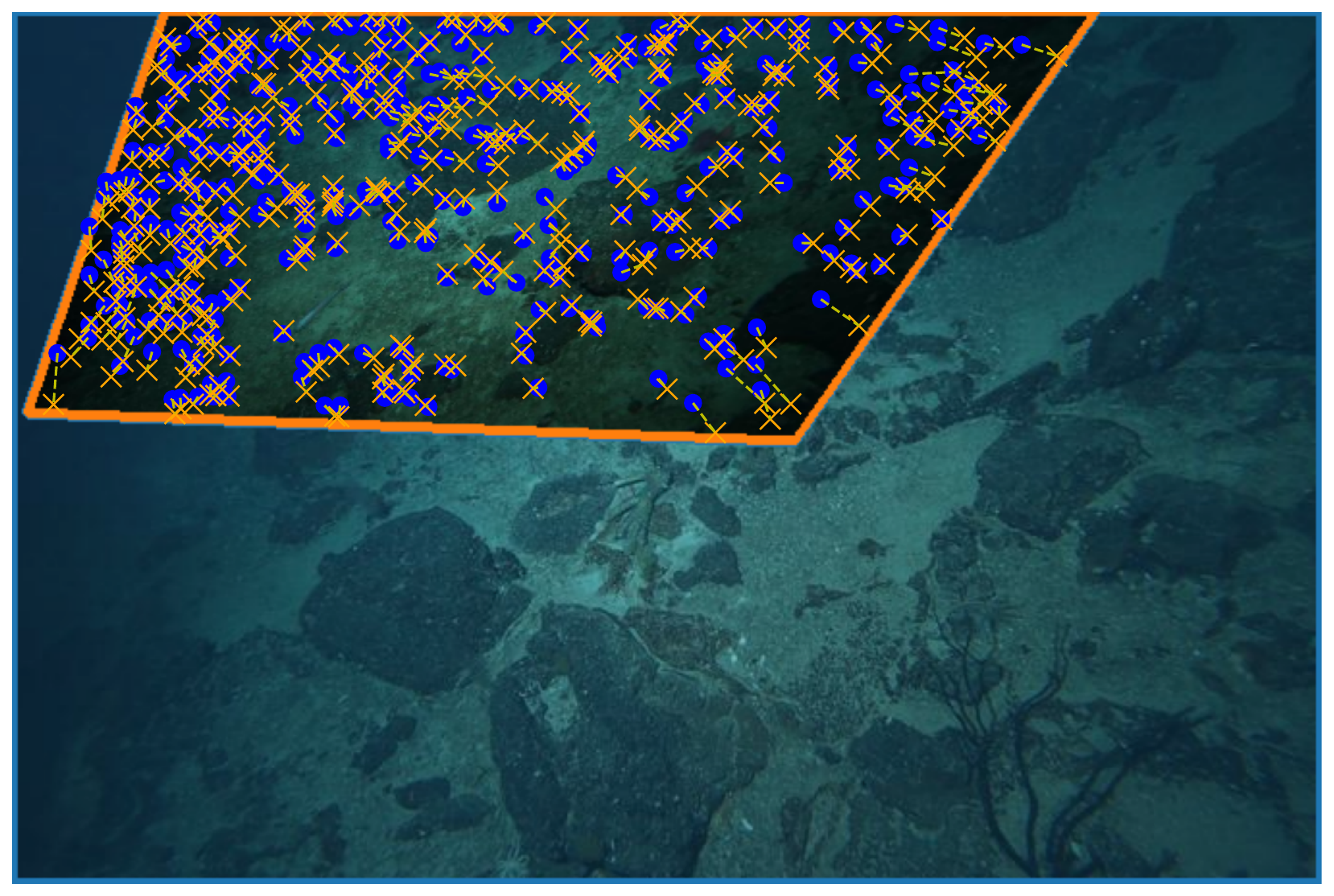} 
&\includegraphics[width=0.50\linewidth]{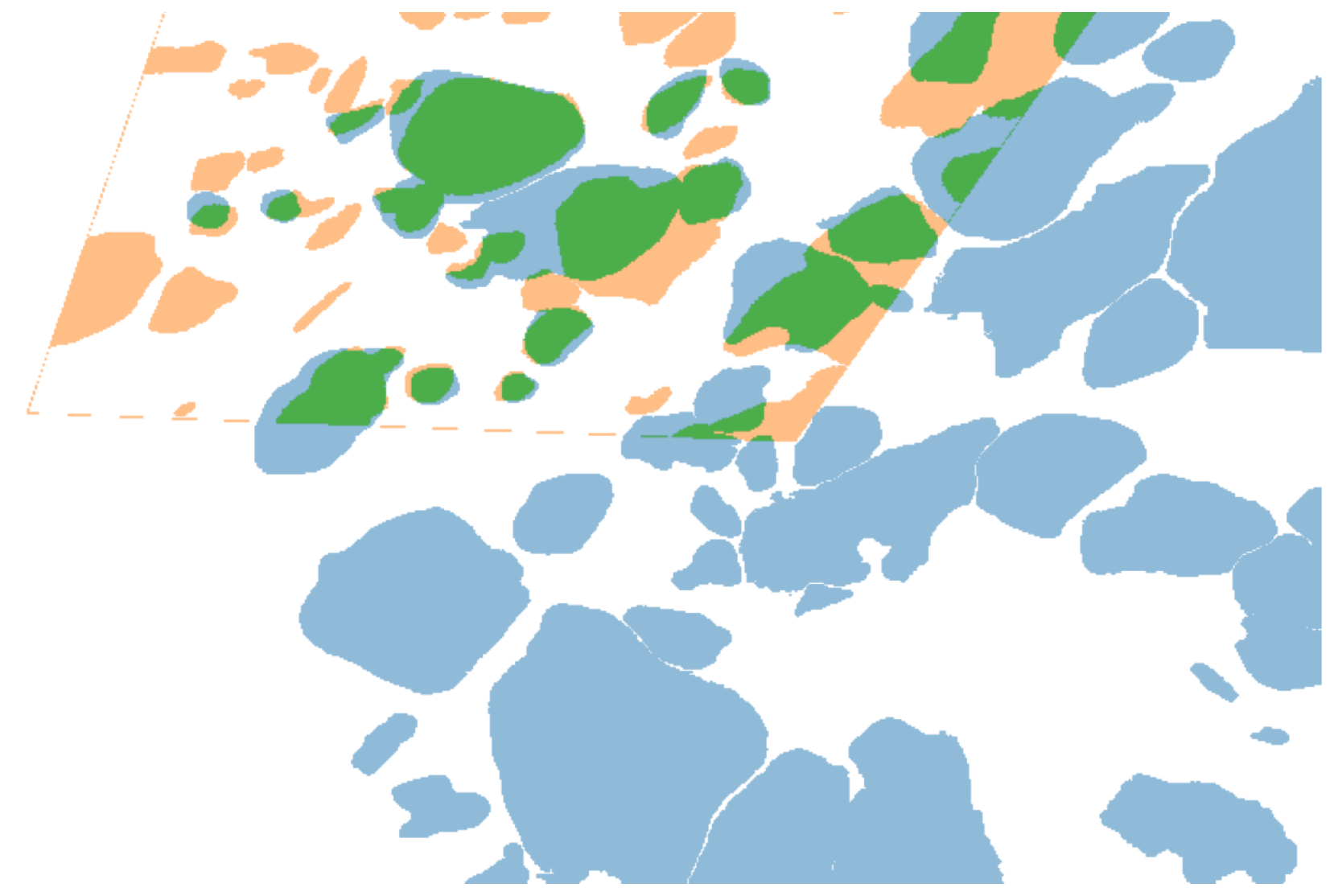}
& {\rotatebox{90}{\LARGE \hspace{1cm}0.45}}
\\
&\includegraphics[width=0.50\linewidth]{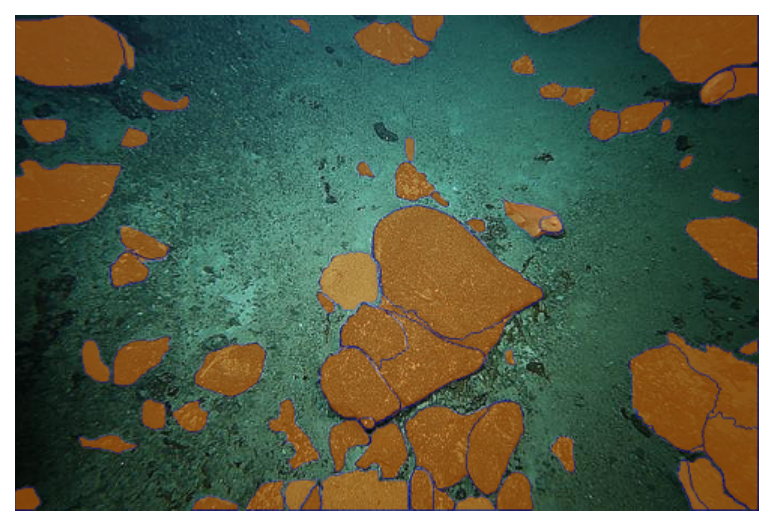}
&\includegraphics[width=0.50\linewidth]{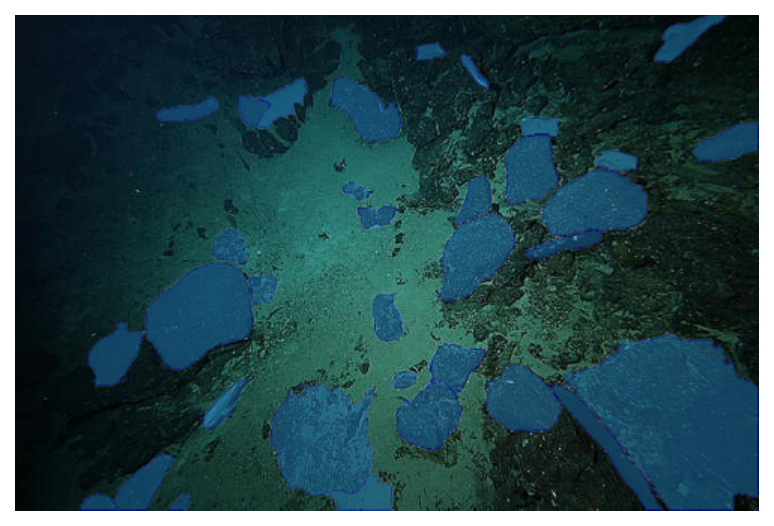} 
&\includegraphics[width=0.50\linewidth]{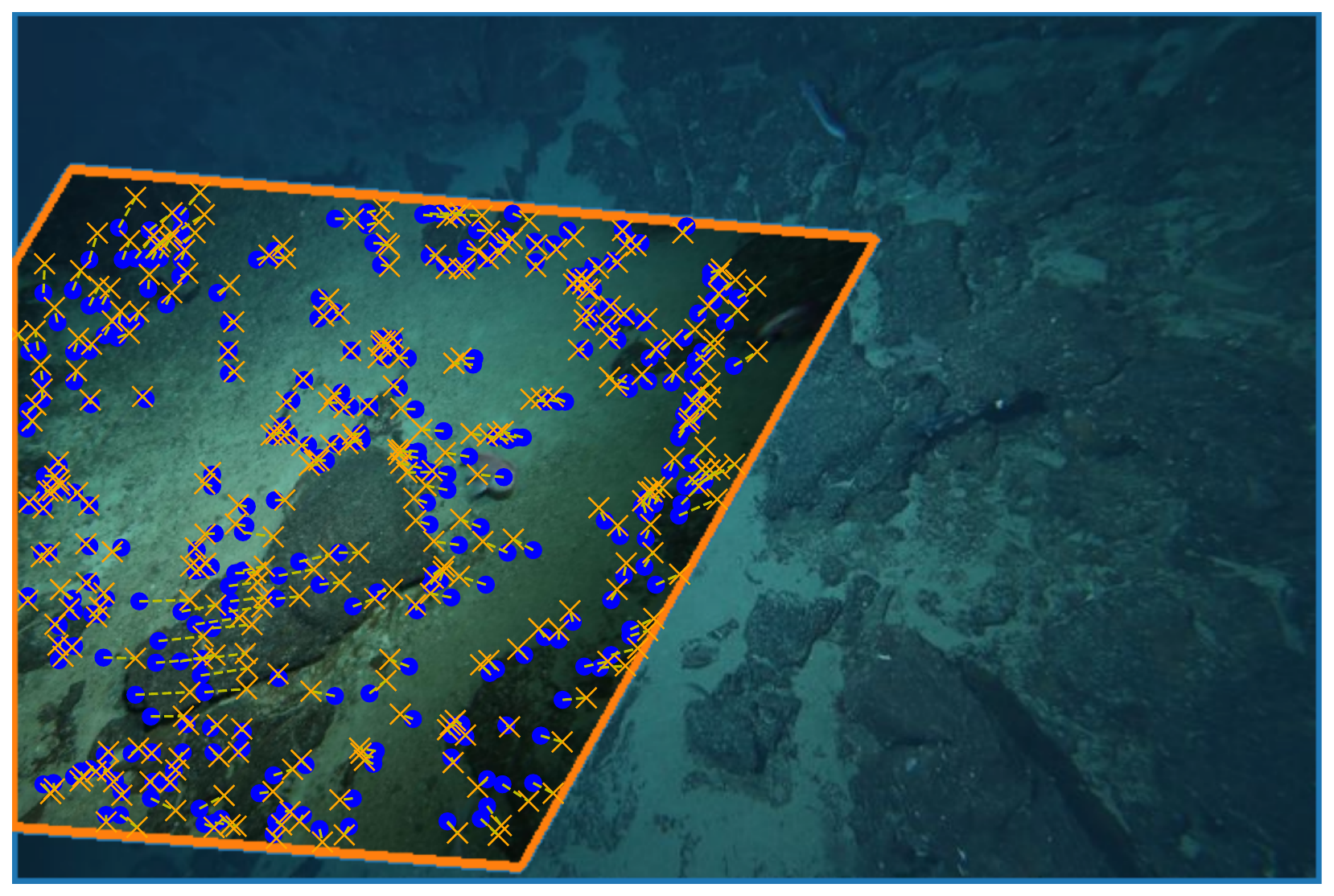} 
&\includegraphics[width=0.50\linewidth]{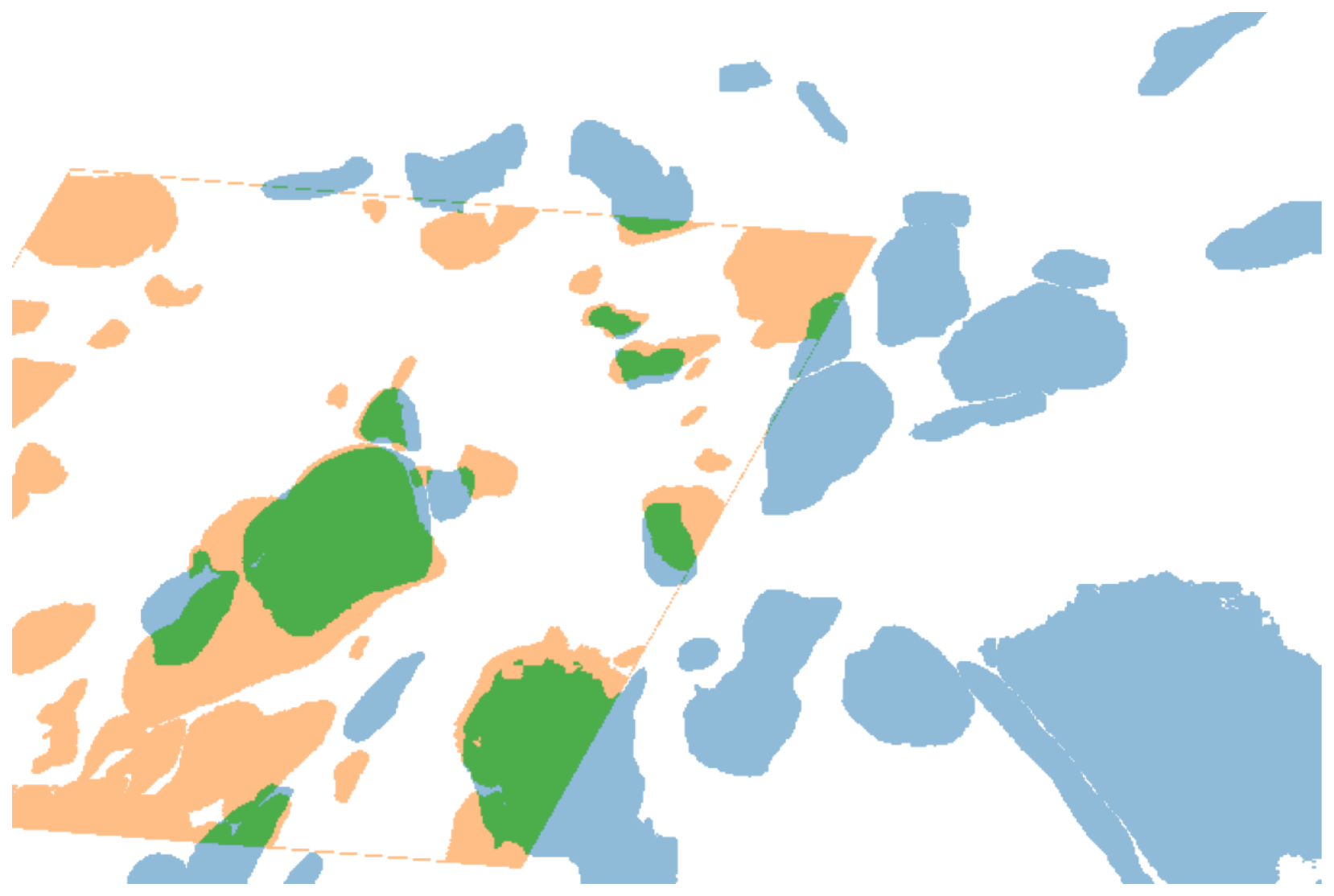}
& {\rotatebox{90}{\LARGE \hspace{1cm}0.33}}
\\
\multirow{2}{*}{\rotatebox{90}{\LARGE St. Helens \hspace{-1cm}}}
&\includegraphics[width=0.50\linewidth]{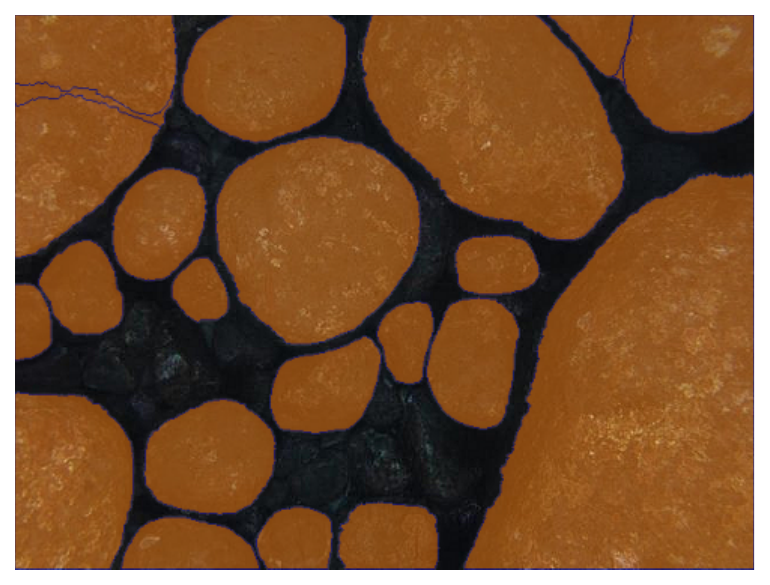}
&\includegraphics[width=0.50\linewidth]{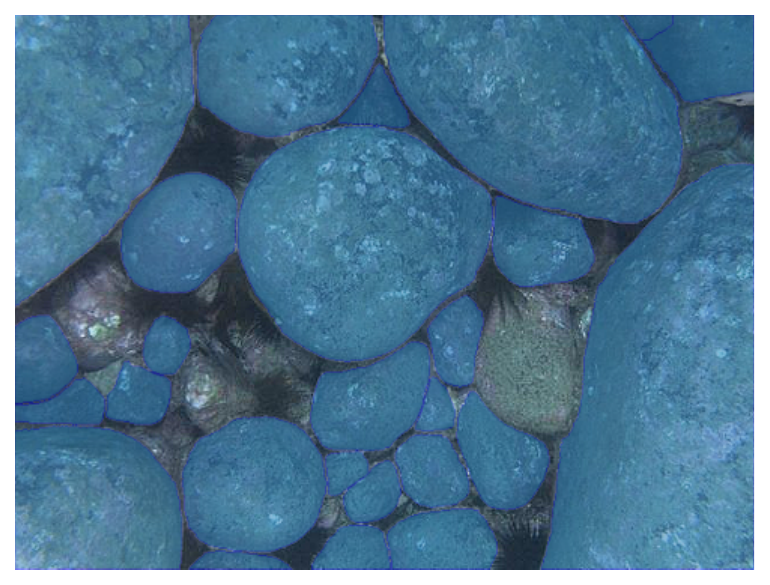} 
&\includegraphics[width=0.50\linewidth]{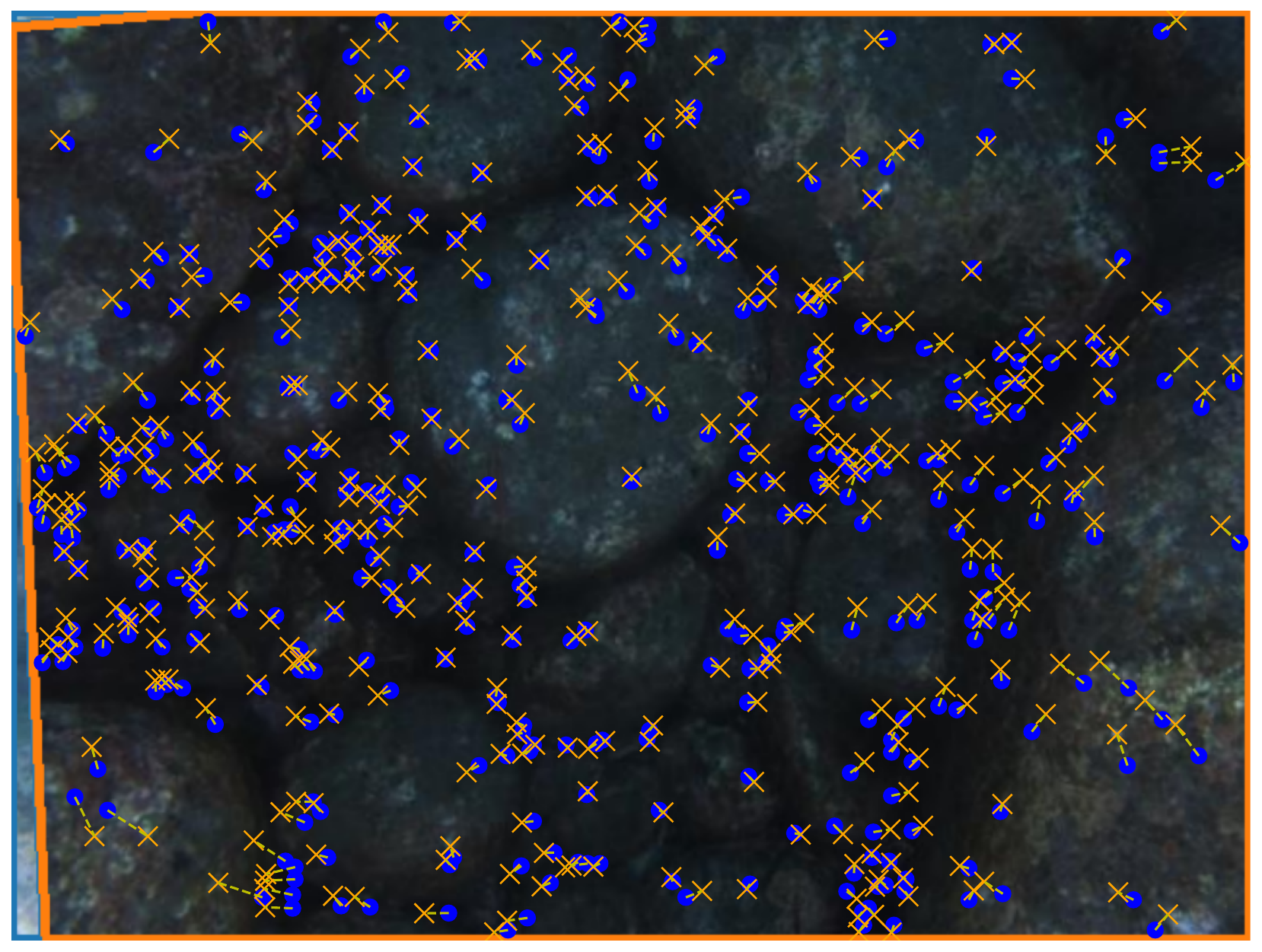} 
&\includegraphics[width=0.50\linewidth]{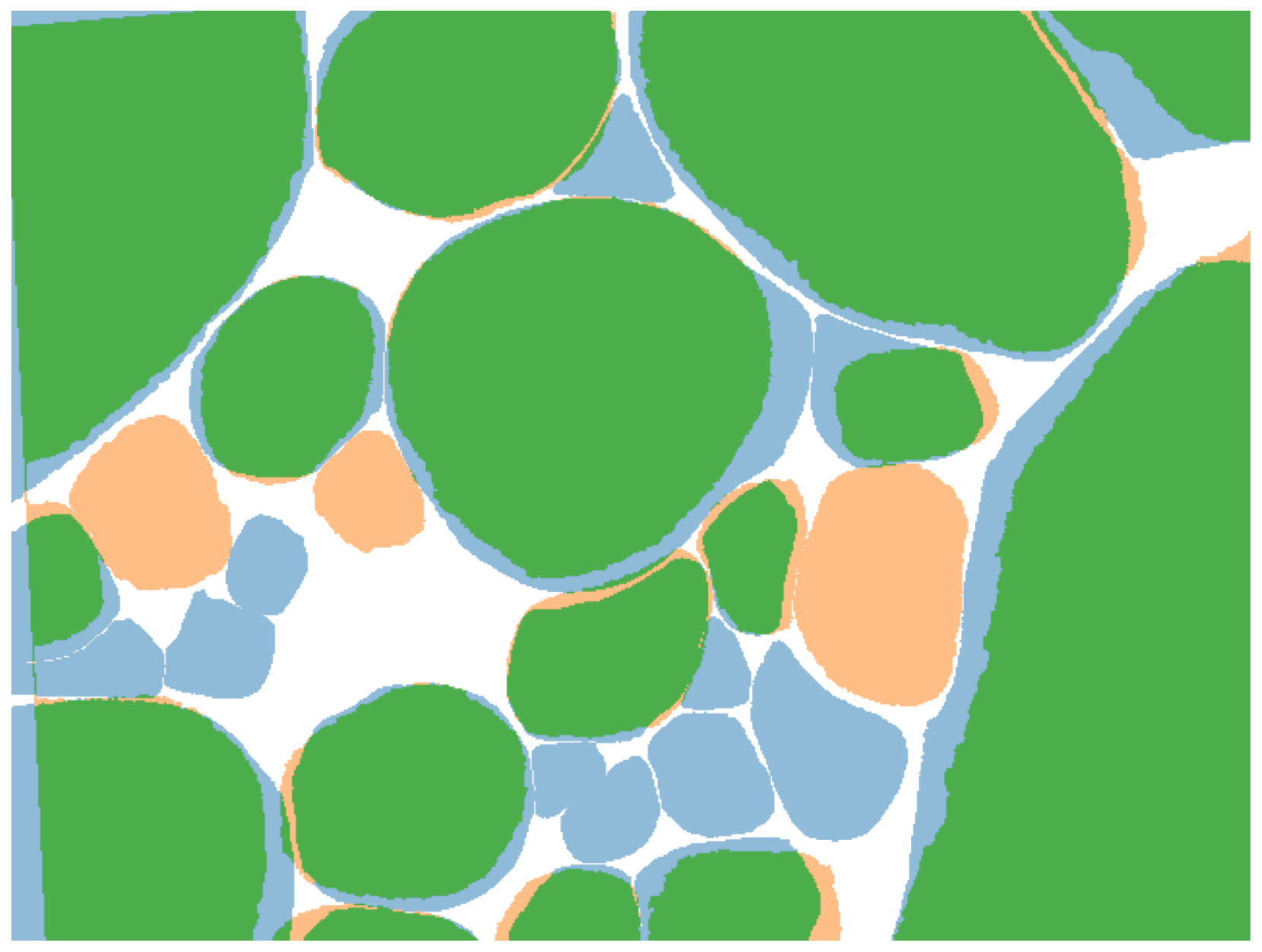}
& {\rotatebox{90}{\LARGE \hspace{1cm}0.83}}
\\
&\includegraphics[width=0.50\linewidth]{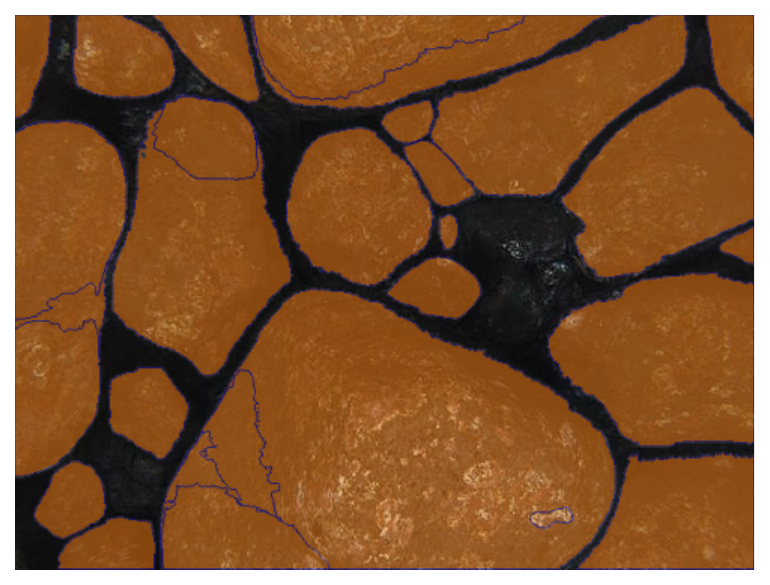}
&\includegraphics[width=0.50\linewidth]{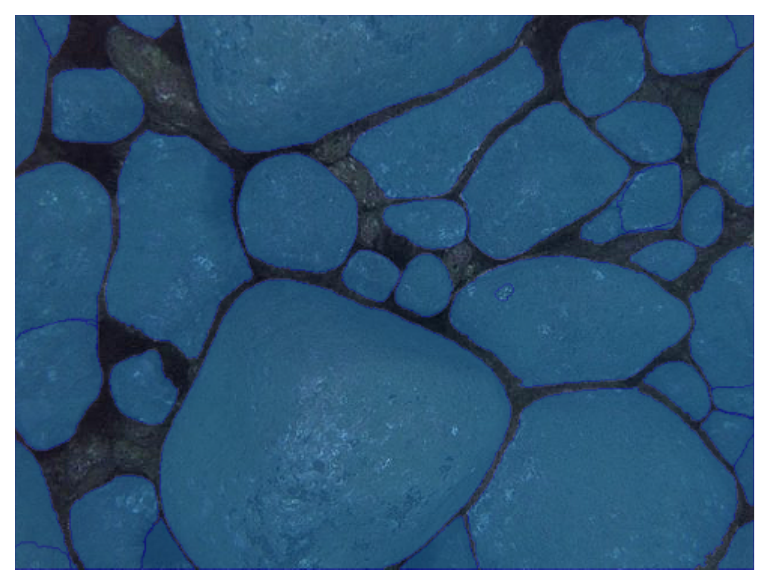} 
&\includegraphics[width=0.50\linewidth]{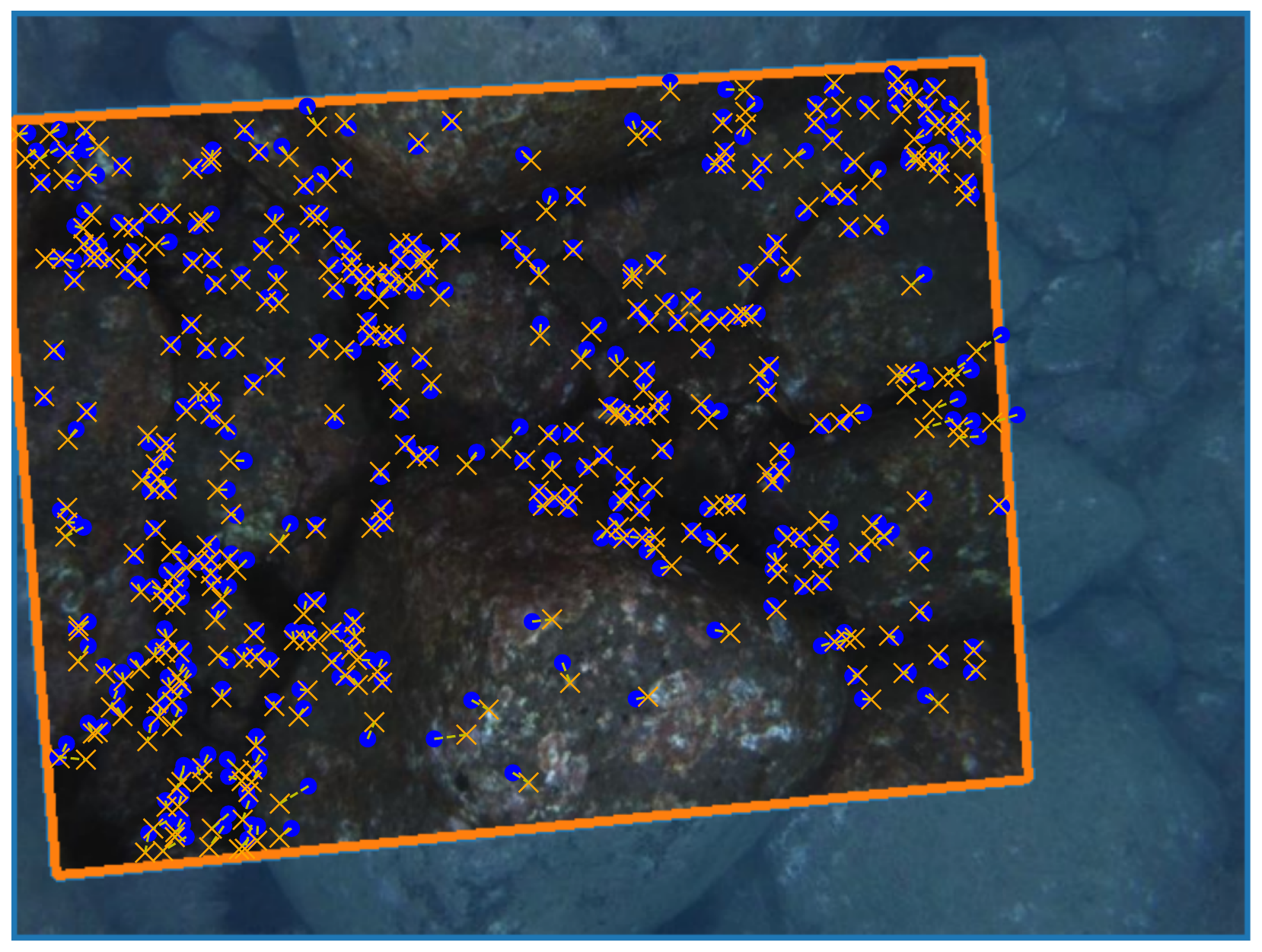} 
&\includegraphics[width=0.50\linewidth]{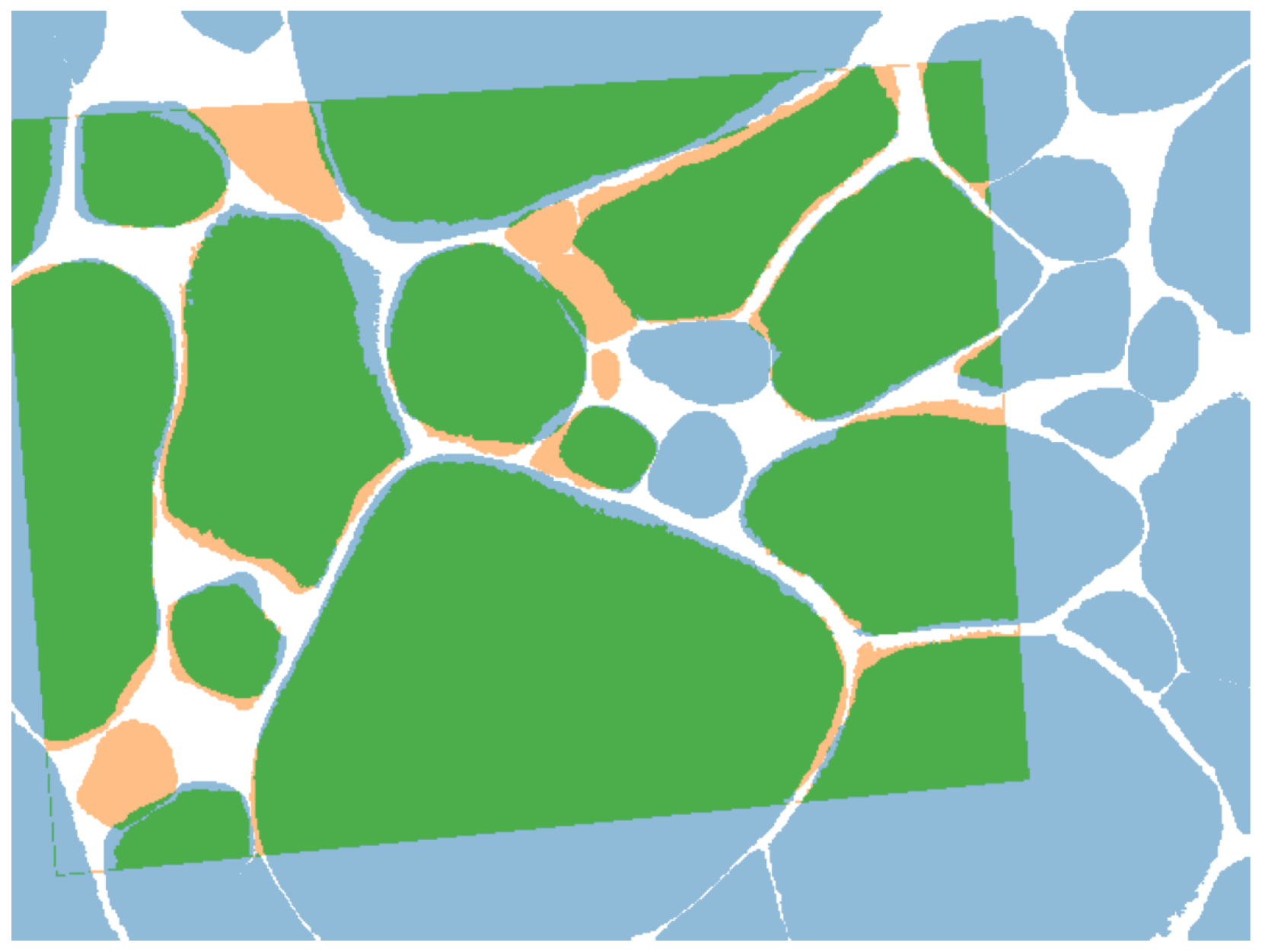}
& {\rotatebox{90}{\LARGE \hspace{1cm}0.86}}
\\
\multicolumn{6}{c}{\includegraphics[]{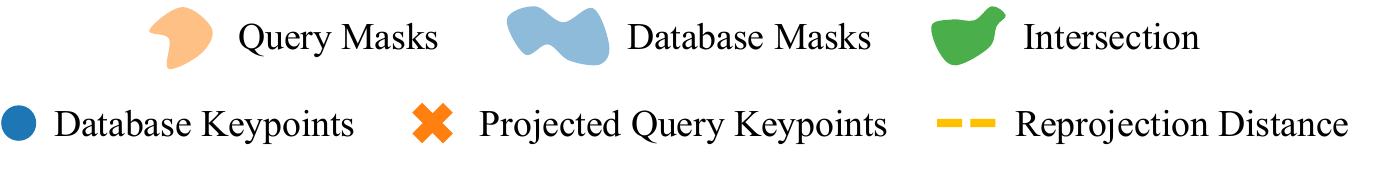}}
\end{tabular}
}
\caption{\textbf{Qualitative results of warping segmentation masks and reprojection error, accompanied by intersection over union (IoU) scores for aligned masks.}  \textbf{Leftmost column:} Query RGB images from each SQUIDLE+ dataset overlaid with SAM2 segmentation masks.  \textbf{Second column:} Database match obtained using our MegaLoc + SuperPoint hierarchical method, overlaid with SAM2 segmentation masks. \textbf{Third column:} Query images warped onto the selected database image using the homography estimated from LightGlue keypoints. {\color{random}Actual keypoints} from the database image are plotted as blue circles, while {\color{mixvpr}projected keypoints} from the query image are shown as orange crosses. The {\color{Dandelion}reprojection distance} is marked with a yellow dashed line. \textbf{Rightmost column:} Query masks warped using the estimated homography and overlaid onto the database masks. Areas of {\color{cosplace}intersection are colored in green}, while non-overlapping query and database masks remain {\color{mixvpr}orange} and {\color{random}blue}, respectively.}
\vspace*{-0.3cm}
\label{fig:segmentation_mre}
\end{figure}

\subsection{Hierarchical Image Retrieval}
\label{sub:hierarchical}
\textbf{Global Retrieval Stage.} Figure~\ref{fig:recall_k} illustrates the Recall@K performance for eight baselines, including the random guesser and brute-force SuperPoint approach (see Section~\ref{sec:vprBaselines}). 

A notable observation in the Okinawa dataset is that for $K \gtrsim 10$, the random guesser's performance begins to approach that of specialized VPR methods. This phenomenon stems from the structured lawnmower trajectory pattern (see Figure~\ref{fig:gps_trajectories}), which creates a high spatial density of images. In such dense patterns, even random selection can occasionally retrieve images within our defined localization radius, despite lacking the visual correspondence that VPR methods identify. This does not indicate an issue with our ground truth, but rather demonstrates that for highly structured survey patterns, the discriminative advantage of VPR methods is most evident at lower $K$ values. The precision-recall curves in Figure~\ref{fig:pr_curves} further illustrate this distinction, showing that VPR methods achieve substantially higher precision than random selection when evaluating the highest-ranked matches.

Table~\ref{tab:vpr_recall} highlights the \textcolor{cosplace}{best Recall@10} results and the \textcolor{orange}{second-best} for each dataset, providing a comparative assessment across methods. MegaLoc is significantly faster than AnyLoc and performs slightly better than CricaVPR on average. We use R@10 for evaluation, as higher values introduce ambiguity due to the limitations of GPS-based ground truth, as discussed above. Due to the significantly higher computational cost of brute-force SuperPoint, particularly when scaled to larger datasets, we exclude this baseline on the St Helens sequence.

\textbf{Local Refinement Stage.} Based on prior experiments, we select MegaLoc as the global image retrieval method in our hierarchical approach. Figure~\ref{fig:pr_curves} shows precision-recall curves for the best single-match results across eight baselines. Our hierarchical method, combining MegaLoc with SuperPoint, achieves performance comparable to brute-force SuperPoint (average precision of \(16\%\) for our hierarchical method vs \(18\%\) for brute-force) while being $100\times$ faster. A key advantage is its ability to first narrow down candidate matches before feature-based refinement, yielding substantial computational savings (see Table~\ref{tab:times}).

The bottom row of Figure~\ref{fig:pr_curves} illustrates the ground truth evaluation, where correctly identified matches are marked in green and false positives appear in red. To ensure reliable evaluation, we threshold with reprojection errors smaller than 10 pixels for a match to be considered valid. This filtering results in precision levels of 39\% for the Eiffel Tower dataset, 99\% for Okinawa, 22\% for Tasman Fracture, and 72\% for St Helens.

To quantify computational efficiency, Table~\ref{tab:times} presents the average computation time per query for each VPR method. By combining MegaLoc with SuperPoint in a hierarchical fashion, our method achieves a $100\times$ speedup over the brute-force SuperPoint approach, demonstrating a favorable trade-off between accuracy and efficiency. All experiments were conducted on an Intel i7 14700K and an NVIDIA RTX 4090 GPU.

\subsection{Segmentation Warping \& Qualitative Results }
Figure~\ref{fig:segmentation_mre} presents the results of warping segmentation masks automatically generated with SAM2 using the homography transformation estimated from LightGlue keypoint correspondences. We evaluate the alignment between the two warped masks using the intersection over union (IoU) metric, where the intersection is defined as the number of shared pixels between the query and database masks, and the union represents the total number of pixels covered by both masks in the warped image.  

This simple evaluation provides a qualitative comparison, demonstrating how a change detection method for underwater ecosystem monitoring can be seamlessly integrated with our image-based registration approach. However, certain limitations affect the accuracy of our change detection: (1) inconsistencies in the segmentation results, and (2) the inability to differentiate between warping inaccuracies, appearance changes, and actual structural modifications in the scene. This is largely due to the reliance on IoU as the sole metric for change detection.

Qualitative observations reveal that scenes with minimal changes, such as the rocky environment in St Helens, achieve higher IoU values. In contrast, significant appearance changes caused by Typhoon Trami in the Okinawa sequence result in lower IoUs, capturing the impact of environmental disturbances. Finally, the Tasman Fracture sequence demonstrates that our approach can register images despite strong viewpoint variations. However, these variations introduce inconsistencies in image segmentation, leading to further reductions in IoU scores, highlighting the potential of combining our approach with more streamlined comparison methods to enable change detection under extreme viewpoint changes. 
\section{Conclusions and Future Work}

We present the first comprehensive benchmark for VPR in underwater environments, addressing a key gap in long-term marine monitoring. Our hierarchical approach, combining global descriptors with local feature refinement, achieves robust performance across diverse underwater scenes over time spans from days to years. Furthermore, precise image registration enables pixel-level comparison between temporally separated observations. This capability is critical for marine conservation, where the vast scale of underwater habitats makes manual assessment impractical.

Future work will focus on fully automating the benchmarking process by leveraging all available resources within SQUIDLE+, enabling the most extensive evaluation of underwater localization pipelines to date.  Equally important is the use of more robust metrics of change detection such as the Structural Similarity Index (SSIM) or semantic definitions of change, such as a shift in class label (e.g., coral to algae) as opposed to persistence (e.g., coral remaining coral). These strategies would help to better isolate biologically significant transformations. We also highlight the potential of semantically aware 3D representations to support spatially grounded change detection and long-term scene understanding that is robust against significant viewpoint changes and non-planar 3D environments. 

\bibliographystyle{IEEEtran}
\bibliography{references/references}

\end{document}